\documentclass[lettersize,journal]{IEEEtran}
\usepackage{amsmath,amsfonts}
\usepackage{algorithmic}
\usepackage{algorithm}
\usepackage{array}
\usepackage[caption=false,font=normalsize,labelfont=sf,textfont=sf]{subfig}
\usepackage{textcomp}
\usepackage{stfloats}
\usepackage{url}
\usepackage{verbatim}
\usepackage{graphicx}
\usepackage{cite}
\usepackage{multirow}
\usepackage{booktabs}
\usepackage{xcolor}
\begin{document}

\title{Prototype-Anchored Generalized Manifold Regression for Unknown-Domain Object Detection}


\author{Zihao~Zhang, Aming~Wu, Yang ~Li and Yahong~Han,~\IEEEmembership{Member,~IEEE.}
        
\thanks{Zihao~Zhang, Yang Li, and Yahong Han are with the College of Intelligence and Computing, Tianjin Key Lab of Machine Learning, Tianjin University, Tianjin 300072, China. E-mail: \{zhangzihao2490, liyang1389, yahong\}@tju.edu.cn. Aming Wu is with the School of Computer Science and Information Engineering, Hefei University of Technology, China. E-mail: amwu@hfut.edu.cn.}}%
\markboth{Journal of \LaTeX\ Class Files,~Vol.~14, No.~8, August~2021}%
{Shell \MakeLowercase{\textit{et al.}}: A Sample Article Using IEEEtran.cls for IEEE Journals}


\maketitle

\begin{abstract}
In this paper, we focus on Single-Domain Generalized Object Detection (Single-DGOD), aiming to transfer a detector trained on one source domain to multiple unknown domains. Existing methods typically rely on simulation-driven paradigms, such as discrete augmentation or static textual prompts, to expand the boundaries of the training distribution. However, finite simulations often fail to capture the infinite dynamic variations of real-world scenarios, which can lead to overfitting on synthetic styles and limit the model's ability to handle complex structural degradations. 
Inspired by the manifold hypothesis, we argue that despite diverse visual variations, semantic features inherently reside on a compact, low-dimensional manifold. Thus, the key to generalization lies in learning to rectify deviant samples back onto this stable manifold, rather than merely exhausting external perturbations.
To this end, we propose a new framework, i.e., Manifold Regression with Visual-Text Dual Chain-of-Thought (MR-DCoT), which reformulates robust generalization as a manifold regression problem. Specifically, we design a Visual-Text Dual Chain-of-Thought module that couples VLM-guided global semantic evolution with diffusion-based local structural perturbations to generate structured off-manifold hard examples. Subsequently, a Class-Specific Prototype Anchoring mechanism is introduced to learn a robust rectification operator that guides deviant features back toward the source semantic manifold. By establishing a closed loop of simulation for outlier generation and regression for semantic correction, our method effectively bridges the distribution gap, significantly boosting generalization and robustness to unseen shifts. Extensive evaluations on three complementary benchmarks, covering diverse driving weather conditions, real-to-art generalization, and zero-shot semantic segmentation, demonstrate the superiority and versatility of our method in handling complex domain shifts.
\end{abstract}    
\vspace{-5pt}
\section{Introduction}
\label{sec:intro}

\IEEEPARstart{S}{ingle}-domain Generalized Object Detection (Single-DGOD) \cite{C_Cap, Div, UFR} represents a challenging yet critical task in computer vision. Its objective is to train a detector using only single-source domain data that can robustly adapt to distribution shifts across multiple unseen target domains. However, this task presents significant challenges. First, the simultaneous requirements for localization accuracy and semantic recognition render object detection highly sensitive to environmental degradations, including local texture variations, abrupt illumination changes, and background interference \cite{domainshift, TIB}. Second, due to the unobservable distribution gap between the source and unknown domains, traditional feature alignment methods frequently prove ineffective \cite{wu2-domain, wu1-universal}.

Existing Single-DGOD methods predominantly follow a Simulation-Driven paradigm: they employ discrete data augmentation \cite{UFR}, style transfer \cite{Poda}, or static textual prompts generated by Vision-Language Models (VLMs) \cite{SECOT, C_Cap, pdoc} to simulate potential target domains, thereby expanding the boundaries of the training distribution. However, this strategy faces an intrinsic limitation: finite simulation can hardly cover infinite real-world variations. Domain shifts in real-world scenarios (e.g., from clear to stormy weather, or from realistic to artistic styles) often exhibit multi-factor coupling, spatiotemporal gradience, and structural degradation \cite{wu2025-OOD2, domainlianxu}. Simple static perturbations or singular textual prompts not only fail to capture these continuous and complex dynamic changes but also risk overfitting the model to limited synthetic styles, leading to performance collapse when encountering genuine unknown domains \cite{DFDD}. 
As discussed in \cite{back}, real-world data in a high-dimensional observation space is typically sparse yet supported by a compact low-dimensional structure: clean samples lie close to this support, whereas corruptions or perturbations tend to push them away from it, creating deviations from the underlying manifold. Building on this perspective, we argue that robust generalization should shift from blindly traversing an effectively infinite perturbation space to mastering a geometric correction rule. Specifically, this requires learning a rectification mechanism that guides off-manifold samples toward a stable semantic manifold.

To address these challenges, we rethink the Single-DGOD problem from the perspective of Regression rather than Simulation. We posit that the essence of robust generalization lies not merely in seeing more diverse samples, but in possessing a stable error-correction capability. We conceptualize source-domain data as distributed on a compact, semantically consistent class-semantic manifold, while unknown-domain samples are modeled as off-manifold points deviated by strong appearance perturbations. Consequently, the key to the Single-DGOD task is not to exhaust all possible perturbation types, but to learn a robust manifold regression mechanism, an ability to stably rectify arbitrary off-manifold inputs back to the neighborhood of the source semantic manifold, thereby locking the semantic essence and geometric structure of targets despite drastic appearance and style changes.

\begin{figure*}[t]
  \centering
  \includegraphics[width=0.9\linewidth]{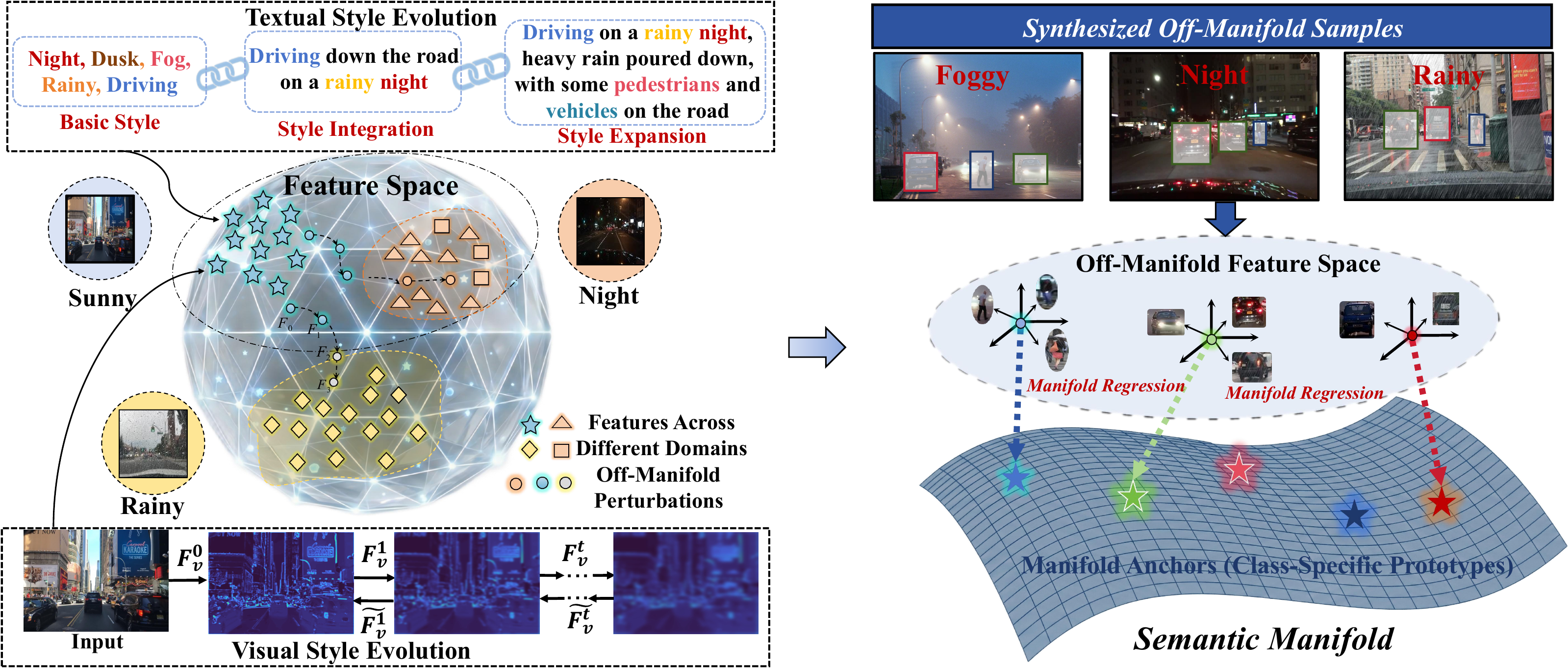}
  \caption{\textbf{Manifold Regression with Visual-Text Dual Chain-of-Thought (MR-DCoT) for detecting unknown-domain objects.} The core of MR-DCoT lies in the duality of simulation and rectification: (a) Off-Manifold Generation employs a Visual-Text Dual Chain-of-Thought to simulate structured off-manifold outliers, driving features away from the source distribution; (b) Manifold Regression learns a robust rectification mechanism that guides these deviant samples back toward the source semantic manifold via prototype anchoring, ensuring semantic consistency against domain shifts.}
    \label{F1}
    \vspace{-10pt}
\end{figure*}

To realize this mechanism, the training process must actively construct sufficiently realistic and structured off-manifold samples as proxies for unseen domain shifts, while simultaneously providing clear regression guidance. To this end, as illustrated in Fig. \ref{F1}, we propose Manifold Regression with Visual-Text Dual Chain-of-Thought (MR-DCoT). The framework consists of two complementary stages: Off-Manifold Generation (depicted in the left part of Fig. \ref{F1}) and Manifold Regression (depicted in the right part of Fig. \ref{F1}). 
MR-DCoT substantially extends our preliminary conference version, SE-COT~\cite{SECOT}. 
While SE-COT mainly employs VLM-generated textual chains to simulate source-domain style variations, MR-DCoT further introduces a diffusion-based visual chain for fine-grained structural perturbation and a prototype-anchored manifold regression mechanism for feature rectification. 
Therefore, MR-DCoT moves beyond simulation-driven style expansion and establishes a closed loop of off-manifold simulation and manifold rectification for unknown-domain generalization.

In the Off-Manifold Generation stage, as depicted in the left panel of Fig. \ref{F1}, we design the Visual-Text Dual Chain-of-Thought (Dual-CoT) module to overcome the limitations of single-modality simulation. Unlike traditional random augmentation, Dual-CoT constructs two synergistic evolutionary chains: the Textual Chain utilizes hierarchical Chain-of-Thought (CoT) generated by Vision-Language Models (VLMs) to specify controllable global semantic directions (e.g., evolving from ‘sunny’ to ‘driving on a rainy night’) \cite{SECOT}; meanwhile, the Visual Chain introduces progressive perturbations based on diffusion concepts \cite{fd}, injecting fine-grained structural noise into the feature space. This mechanism, where language sets the high-level direction and vision fills in fine-grained details, generates off-manifold samples with large-magnitude shifts while preserving semantic invariance, thereby serving as effective proxies for unknown domains.

In the subsequent manifold regression stage, capitalizing on these structured off-manifold proxies (visualized as Synthesized Off-Manifold Samples in the right panel of Fig. \ref{F1}), we formulate generalization as an optimizable class-conditional rectification objective. We introduce a Class-Specific Prototype Anchoring mechanism, where prototypes learned from the source domain serve as stable manifold anchors on the underlying semantic manifold surface. 
For the off-manifold features generated by Dual-CoT, we impose a prototype-anchored rectification objective that guides them toward the corresponding class-conditional prototype neighborhoods.
As shown by the rectification trajectories in Fig. \ref{F1}, this process encourages the network to disregard the injected stylistic and structural perturbations, learning a robust rectification operator that guides deviant features back to the semantic manifold. Consequently, this establishes a closed loop of Simulate-to-Deviate, Regress-to-Rectify, enabling the model to lock onto semantic identity regardless of severe domain shifts.

Significant performance improvements across diverse adverse weather scenarios, Real-to-Art generalization, and zero-shot semantic segmentation benchmarks demonstrate the superiority of MR-DCoT, validating the effectiveness of the manifold paradigm: Simulation for generating outliers, Regression for returning to the semantic manifold.

To summarize, our contributions are as follows:

• We redefine Single-DGOD as a manifold regression problem, shifting from simulation coverage to learning a stable error-correction capability for off-manifold perturbations.

• We propose MR-DCoT, which utilizes a Visual-Text Dual Chain-of-Thought to construct structured off-manifold proxies, effectively synergizing global semantic evolution with local structural perturbations.

• We establish a Simulate-to-Deviate, Regress-to-Rectify closed loop via Class-Specific Prototype Anchoring, optimizing a robust rectification operator that anchors deviant features toward manifold-consistent semantic neighborhoods.
\section{Related Work}
\label{sec:related}

\subsection{Single-Domain Generalized Object Detection}
Single-Domain Generalized Object Detection (Single-DGOD) \cite{C_Cap, S_DG} aims to train a robust detector using only a single source domain that can generalize to arbitrary unseen target domains. Existing methodologies can be broadly categorized into two paradigms: Domain-Invariant Feature Learning and Data Augmentation.
In the Domain-Invariant Feature paradigm \cite{pamidoj}, methods like S-DGOD \cite{S_DG} introduce a cyclic-disentangled self-distillation framework to isolate invariant features, while FWCL \cite{FWCL} combines frequency whitening with contrastive learning to suppress domain-sensitive frequency components. Although these approaches improve robustness, their efficacy heavily relies on the assumption that style and content can be perfectly decoupled, an assumption that often falters under severe distribution shifts, such as heavy rain or dense fog, where semantic structures are degraded. 
The Data Augmentation and Diversification paradigm \cite{sfda, Div} focuses on expanding the training distribution to simulate potential domain shifts. For instance, UFR \cite{UFR} proposes an unbiased training strategy via multi-view augmentation to mitigate source domain overfitting; DIV \cite{Div} introduces a diversification mechanism to generate random style perturbations; and G-NAS \cite{G-NAS} leverages neural architecture search to discover network structures with intrinsic robustness. However, these methods typically rely on discrete or random augmentation techniques, failing to capture the continuous and structured nature of real-world environmental changes.
To address this, we propose a Manifold Regression paradigm, shifting focus from simulation expansion to stable error-correction.

\subsection{Vision-Language Models for Domain Generalization}
The advent of Vision-Language Models (VLMs) like CLIP \cite{CLIP} has revolutionized domain generalization by providing rich, open-set semantic priors. A common paradigm involves using natural language prompts to guide the synthesis of out-of-distribution samples.
Methods like \cite{C_Cap} and \cite{Poda} utilize one-step prompts (e.g., ‘Driving on a rainy night’) to hallucinate target styles, leveraging the cross-modal alignment of CLIP to enforce semantic consistency in the generated samples. More recently, SE-COT\cite{SECOT} introduced a Chain-of-Thought strategy to progressively evolve text prompts, simulating a sequence of style changes rather than a single static shift.
However, a critical limitation remains: these approaches are predominantly text-driven. Relying solely on language limits the simulation to global style transfers (e.g., modifying global appearance statistics such as illumination tone or color distribution) but often neglects fine-grained visual structural degradations (e.g., local blurring, noise patterns, or edge corruption) that are difficult to describe linguistically. Consequently, the generated samples may lack the structural complexity of real-world domain shifts.
 To bridge the gap between semantic control and visual detail, we introduce a Visual-Text Dual Chain-of-Thought (Dual-CoT). By coupling hierarchical textual evolution with diffusion-based visual perturbations \cite{DGS}, our method simultaneously models global semantic progression and local structural degradation, generating outliers that are both semantically diverse and structurally challenging.

\subsection{Manifold Learning}
The concept of regression on manifolds has distinct roots in statistical learning, where geodesic regression is employed to model data on Riemannian surfaces~\cite{Georeg}. In the context of deep representation learning, a cornerstone is the manifold hypothesis, which posits that high-dimensional visual data reside on compact, low-dimensional structures. Recently, Li et al.~\cite{back} reinforced this perspective through a denoising lens, demonstrating that clean data concentrate near a low-dimensional support while perturbations drive samples off-manifold. Crucially, they proved that directly predicting on-manifold data ($x$-prediction) is fundamentally more robust than modeling off-manifold quantities. This insight aligns with deep-seated theories: Contractive Auto-Encoders (CAE)~\cite{CAE} enforce feature contraction toward the manifold via Jacobian regularization to achieve geometric invariance, while Denoising Score Matching~\cite{ScoreMatching} mathematically defines the correction process as learning a vector field pointing toward the high-density manifold. While these generative approaches implicitly perform manifold rectification by regressing noisy inputs to their clean counterparts, we explicitly tailor this paradigm for Single-DGOD. Different from manifold-valued regression or generative score fields, we instantiate rectification in a discriminative detection feature space with class-conditional anchors and paired instance-level regression. Ultimately, we advocate for a shift from traversing infinite perturbations to mastering a geometric correction rule: learning a parametric operator to rectify deviant samples toward the stable semantic manifold to ensure robust generalization.
\section{Method}
\begin{figure*}[!t]
  \centering
  \includegraphics[width=2.0\columnwidth]{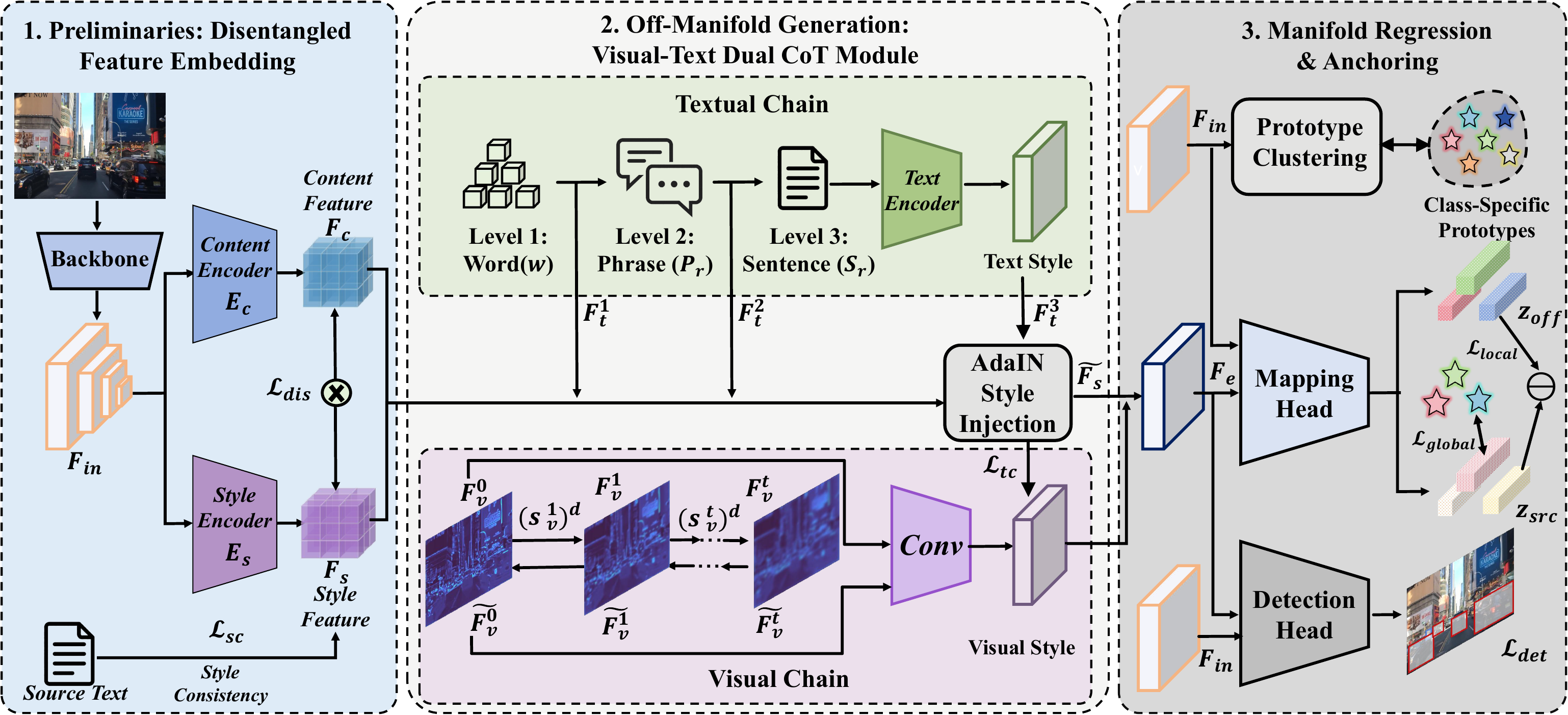}
  \caption{\textbf{Overview of the Manifold Regression with Visual-Text Dual Chain-of-Thought (MR-DCoT) framework.} First, the input image features are decoupled into style ($F_s$) and content ($F_c$) components via the Disentangled Feature Embedding module, ensuring semantic integrity through contrastive and consistency losses. Next, the Visual-Text Dual-CoT Module is employed to generate structured off-manifold outliers. The Textual Chain evolves hierarchical semantic descriptions (Word $\to$ Phrase $\to$ Sentence) to guide global style directions, while the Visual Chain injects fine-grained structural perturbations to simulate continuous feature evolution. Then, the evolved style and distorted content are integrated via AdaIN Style Injection to synthesize hard off-manifold examples. 
  Finally, Manifold Regression via anchoring is applied to rectify these deviant features toward the source semantic manifold by guiding them to the corresponding class-conditional prototype neighborhoods, ensuring robust generalization against unseen domain shifts.
    }
    \label{F2}
     \vspace{-0.1in}
\end{figure*}

As illustrated in Fig.~\ref{F2}, we tackle Single-Domain Generalized Object Detection (Single-DGOD) from a manifold perspective, and propose MR-DCoT, i.e., Manifold Regression with Visual--Text Dual Chain-of-Thought. 
By reformulating robust generalization as a manifold regression problem, MR-DCoT learns a stable parameterized rectification mechanism that maps off-manifold samples induced by unseen domain shifts back toward the semantic manifold of the source domain. 
In this work, the semantic manifold refers to the class-conditional semantic support formed by source-domain ROI-level instance features in the detector feature space.
We further approximate each class-conditional semantic support with prototype-centered semantic neighborhoods in the subsequent manifold regression stage, as detailed in Sec.~\ref {sec3.3}.
The framework consists of three key components: Disentangled Feature Embedding, a Visual--Text Dual Chain-of-Thought module for structured outlier simulation, and manifold regression via class-specific prototype anchoring. 
MR-DCoT is a substantial extension of our preliminary conference version SE-COT~\cite{SECOT}, moving from text-driven style simulation to a closed loop of visual-text off-manifold simulation and prototype-anchored feature rectification, with detailed comparisons provided in Sec.~III-D.
The overall training procedure of the proposed MR-DCoT framework is summarized in Algorithm~\ref{alg:mrdcot}.

\vspace{-5pt}
\subsection{Preliminaries: Disentangled Feature Embedding}
To simulate effective domain shifts without compromising semantic integrity, we begin by disentangling the backbone features from the first layer into distinct style and content components. Specifically, we employ two lightweight encoders, $E_s$ and $E_c$, to decouple the input feature map $F_{in}$ into style features $F_s$ and content features $F_c$:
\begin{equation}
F_s = E_s(F_{in}), \quad F_c = E_c(F_{in}).
\end{equation}

To encourage effective separation and reduce information overlap between $F_s$ and $F_c$, we impose a decorrelation constraint.
We define $\mathrm{sim}(a,b)$ as the cosine similarity between globally pooled and $\ell_2$-normalized representations. The disentanglement loss is defined as:
\begin{equation}
\mathcal{L}_{dis} = \left|\mathrm{sim}(F_s,F_c)\right|.
\end{equation}
Minimizing this objective encourages the global style and content representations to be approximately orthogonal, thereby suppressing cross-branch correlation and reducing redundant information.
Additionally, we leverage the source text embedding $F_t^{src}$ to calibrate the style branch. To bridge the dimensional gap between visual features and the text embedding space, we introduce a learnable projection head $\psi$. The style consistency loss is defined as:
\begin{equation}
\mathcal{L}_{sc} = 1 - \mathrm{sim}(\psi(\mathrm{Pool}(F_s)), F_t^{src}),
\end{equation}
where $\mathrm{Pool}(\cdot)$ denotes global average pooling, and $F_t^{src}$ is the source-domain text embedding extracted by a pre-trained encoder (e.g., CLIP \cite{CLIP}). Consequently, the total objective for the disentangled embedding stage is formulated as:
\begin{equation}
\mathcal{L}_{emb} = \mathcal{L}_{dis} + \lambda_{sc}\mathcal{L}_{sc},
\end{equation}
where $\lambda_{sc}$ is a hyperparameter controlling the contribution of the style consistency loss, we set $\lambda_{sc} = 0.5$ to balance the orthogonality constraint with the source style prior.

\begin{figure}[t]
  \centering
  \includegraphics[width=1.0\linewidth]{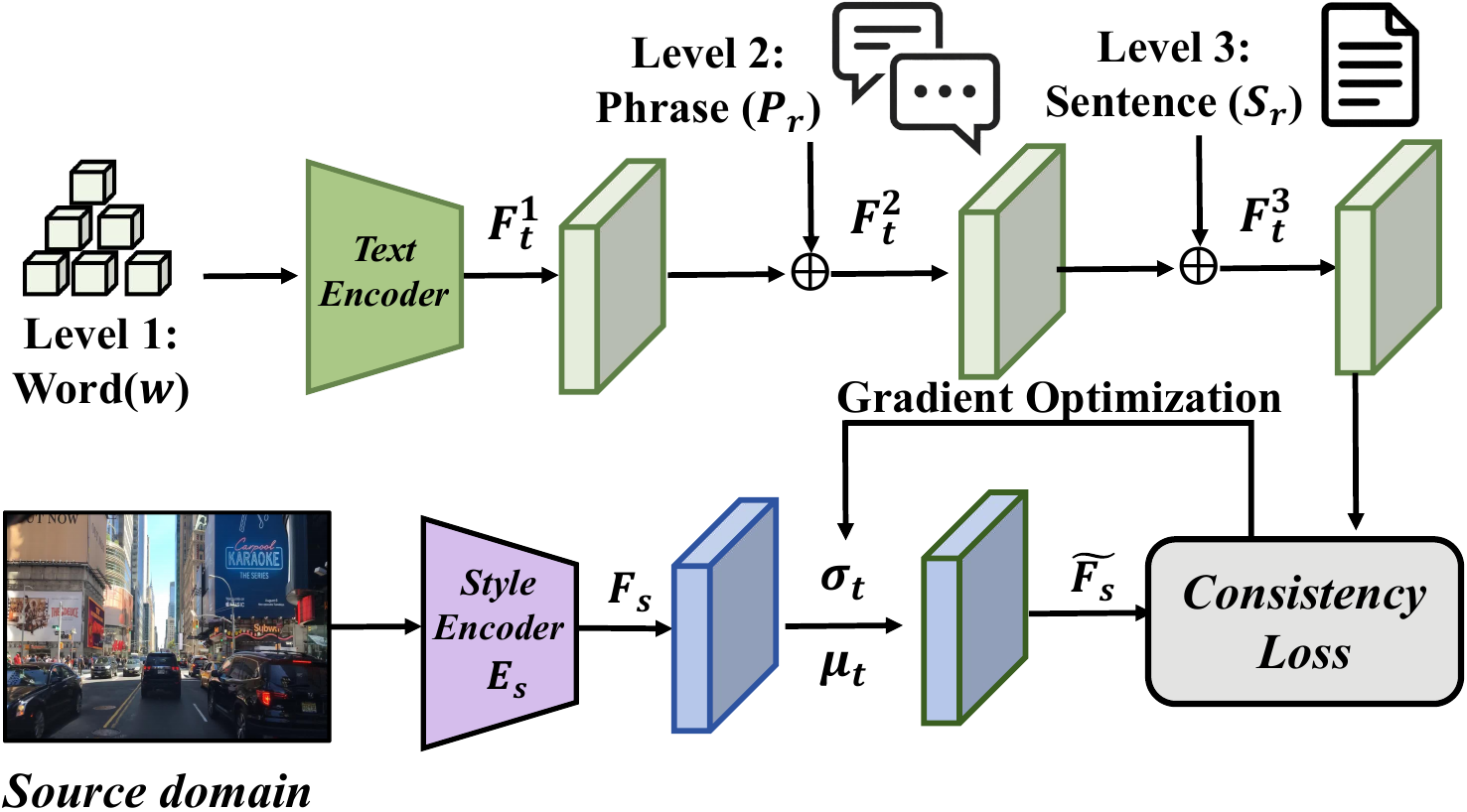}
  \caption{
  \textbf{Complex Style Evolution via Textual Chains.} By employing text prompts that progress from simplicity to complexity, the style is continuously evolved and expanded, thereby simulating a wider variety of style features with distinct data distributions. Parameters $\mu_t$ and $\sigma_t$ are trained using consistency loss to align textual features with visual features.}
  \vspace{-0.2in}
    \label{F3}
\end{figure}

\subsection{Off-Manifold Generation: Visual-Text Dual-CoT Module}
\label{sec:dual_cot}
To construct off-manifold hard examples with larger shift magnitudes and higher diversity, thereby strengthening the supervision for learning a robust regression mechanism, we propose the Visual--Text Dual Chain-of-Thought module (Dual-CoT). Built upon disentangled content and style features, Dual-CoT couples a textual chain that specifies a controllable global semantic direction with a diffusion-based visual chain that generates feature trajectories progressively departing from the source manifold. The textual chain drives global style evolution, while the visual chain injects fine-grained local variations to complement the abstraction of textual descriptions. Unlike random augmentation \cite{Div}, Dual-CoT generates controllable off-manifold trajectories under visual-text semantic guidance, which enlarges the domain shift while preserving category-level semantic consistency, thereby providing structured proxies for subsequent manifold regression.

\begin{algorithm}[tb]
    \caption{Manifold Regression with Visual-Text Dual-CoT}
    \label{alg:mrdcot}
\begin{algorithmic}[1]
    \STATE {\bfseries Input:} Source images $\mathcal{I}$, Ground-truth annotations $\mathcal{Y}$, Hierarchical text prompts $\mathcal{T}$.
    \STATE {\bfseries Output:} Optimized detector parameters $\Theta$, Learned class-specific prototypes $\mathcal{P}$.
    
    \STATE \textit{// Step 1: Disentangled Feature Embedding}
    \STATE Decouple backbone features $F_{in}$ into style $F_s$ and content $F_c$ via $E_s, E_c$ using Eq.~(1).
    \STATE Compute embedding loss $\mathcal{L}_{emb}$ using Eqs.~(2)--(4).
    
    \STATE \textit{// Step 2: Off-Manifold Generation (Dual-CoT)}
     \STATE \textbf{Textual Chain:} Evolve style $\tilde{F}_s$ via hierarchical prompts $F_t^1 \to F_t^3 \in \mathcal{T}$ and AdaIN using Eqs. (5)-(9).
    \STATE \textbf{Visual Chain:} Generate restored-yet-shifted content $\hat{F}_v^0$ via blur-diffusion $\mathcal{B}(\cdot)$ and reverse network $\mathcal{D}$ using Eqs.~(10)--(13).
    \STATE Compute generation consistency losses $\mathcal{L}_{align}$ and $\mathcal{L}_{diff}$ using Eq.~(9) and Eq.~(14).
    \STATE \textbf{Integration:} Synthesize off-manifold feature $F_e = \mathrm{Conv}(\mathrm{Concat}(\tilde{F}_s, \hat{F}_v^0, F_v^0 ))$ using Eq.~(15).

    \STATE \textit{// Step 3: Manifold Regression via Anchoring}
    \STATE Define the source semantic manifold and prototype-centered neighborhoods using Eqs.~(16)--(18).
    \STATE Update class-specific prototypes $\mathcal{P}$ via soft assignment and residual aggregation using Eqs.~(19)--(21).
    \STATE Generate prototype-enhanced feature $F_p$ using Eq.~(22).
    \STATE Extract instance features $z_{src} = \mathrm{ROI}(F_{in})$ and $z_{off} = \mathrm{ROI}(F_e)$ using ground-truth boxes from $\mathcal{Y}$.
    \STATE Map features to the normalized prototype space via the shared mapping head $\mathcal{H}(\cdot)$ and compute $\mathcal{L}_{reg} = \mathcal{L}_{global} + \lambda_{loc} \mathcal{L}_{local}$ using Eqs.~(23)--(25).

    \STATE \textit{// Step 4: Global Optimization}
    \STATE Optimize the detector with both prototype-enhanced source features $F_p$ and off-manifold features $F_e$.
    \STATE Update parameters $\Theta$ by minimizing the aggregated total objective $\mathcal{L}_{total}$ via backpropagation.
    
    \STATE {\bfseries Return} $\Theta, \mathcal{P}$
\end{algorithmic}
\end{algorithm}

\subsubsection{Textual Chain: CoT-Guided Global Style Evolution}
\label{sec:text_chain}
To ensure precise style simulation, we optimize the AdaIN modulation parameters $\mu_t$ and $\sigma_t$ in an independent training phase, where the detector is frozen and only a lightweight text-conditioned projection head is trained to predict $(\mu_t,\sigma_t)$ from text features. 
As illustrated in Fig. \ref{F3}, we construct a hierarchical text generation process to guide this evolution. 
Initially, we employ a refined image captioning model \cite{imagecaption} to extract pertinent keywords (e.g., sunny, realistic, driving scene, natural scene) from source domain images. 

Subsequently, instead of manually defining fixed semantic dimensions such as weather, time, style, action, and detail, we adopt a source-conditioned semantic factor discovery strategy. 
Given the source-domain keywords, GPT \cite{gpt} is prompted to infer several category-preserving domain variation factors and generate corresponding keyword groups. 
These factors describe possible appearance or domain changes, such as illumination, weather, texture, imaging condition, scene context, or structural degradation, without changing the object categories. 
We denote the generated keyword groups as $\mathcal{G}=\{G_j\}_{j=1}^{M}$, where $G_j=\{w_j^1,w_j^2,\ldots,w_j^{m_j}\}$ is the $j$-th keyword group.

During training, we randomly sample one keyword group $G_r$ from $\mathcal{G}$ and use CLIP to extract features from all keywords in this group. 
To improve reproducibility and avoid instability from stochastic language generation, GPT is only used offline, and all generated keyword groups are fixed throughout training and evaluation without using any target-domain images, annotations, or statistics.

We utilize CLIP to extract features from these words and perform initial fusion to obtain the Level 1 textual features $F_t^1$:
\begin{equation}
F_t^1 = \frac{1}{m_r}\sum_{i=1}^{m_r} E_{text}(w_r^i),
\end{equation}
where $m_r=|G_r|$ denotes the number of keywords in the sampled group. 
In this way, the textual chain no longer depends on task-specific, predefined vocabularies.

Building upon this, we employ GPT \cite{gpt} to synthesize these discrete words into coherent phrases $P_r$ (e.g., `Driving down the road on a rainy night'), thereby integrating local context. 
The features extracted from these phrases are aggregated with the previous level's features:
\begin{equation}
F_t^2 = E_{text}(P_r) + F_t^1.
\end{equation}

In the final stage of the textual chain, we expand the phrases into complete sentences $S_r$ by incorporating detailed stylistic descriptions. 
The sentence generation is also conditioned on the sampled keyword group $G_r$ and follows the same category-preserving constraint, ensuring that the generated descriptions only describe domain variations rather than changing object categories. 

This step simulates a wider variety of data distributions and complex real-world scenarios. 
The sentence-level features are aggregated with $F_t^2$ to yield the final refined textual guidance $F_t^3$:
\begin{equation}
F_t^3 = E_{text}(S_r) + F_t^2.
\end{equation}

To leverage these textual features for visual style evolution, we adopt an AdaIN-based modulation scheme \cite{ada} that injects text-driven style statistics into the normalized source feature. 
Specifically, given the source style feature map $F_s$, we first remove its original channel-wise statistics by normalization, and then re-scale and shift the normalized features using text-predicted style statistics to obtain the evolved style feature:
\begin{equation}
\tilde{F}_s = \sigma_t \odot \frac{F_s-\mu(F_s)}{\sigma(F_s)+\epsilon}+\mu_t,
\end{equation}
where $\mu(F_s)$ and $\sigma(F_s)$ are the channel-wise mean and standard deviation of $F_s$, respectively, $\odot$ denotes channel-wise element-wise multiplication, and $\epsilon$ is a small constant for numerical stability. 
The modulation parameters $(\mu_t,\sigma_t)$ are conditioned on the refined textual guidance $F_t^3$, where $\sigma_t$ controls the channel-wise scaling and $\mu_t$ provides the channel-wise bias, thereby explicitly imprinting the target style semantics onto the visual feature distribution.
All textual components are generated offline from source-conditioned descriptions and fixed before training, ensuring that the final textual guidance $F_t^3$ is reproducible without querying LLMs or using any target-domain images or annotations during training or inference.

To guarantee semantic consistency between the evolved visual style and the hierarchical textual guidance, we enforce a style alignment constraint defined as:
\begin{equation}
\mathcal{L}_{align} = 1 - \mathrm{sim}(\psi(\mathrm{Pool}(\tilde{F}_s)), F_t^3),
\end{equation}
where $\text{Pool}(\cdot)$ and $\psi$ are consistent with the definitions in Eq. (3), ensuring the visual features are properly mapped into the shared embedding space. 
Minimizing this objective explicitly optimizes the modulation parameters $(\mu_t, \sigma_t)$ to capture the target stylistic semantics.

\subsubsection{Visual Chain: Diffusion-Guided Local Feature Evolution}  
Inspired by diffusion models \cite{Diffs, tpamidif} and UCFD \cite{DFDD}, we construct a progressive feature diffusion process to guide local feature evolution. Unlike standard forward diffusion that directly injects Gaussian noise, we employ a Gaussian blur as the primary corruption operator to better preserve geometric and semantic structures. In our framework, the diffusion is performed on the disentangled content feature $F_c$, which is designated as the initial diffusion state $F_{v}^{0}$.

\noindent\textbf{Feature Forward Diffusion.}
Let $\mathcal{B}(\cdot, s_v^{t})$ denote a Gaussian-blur diffusion operator, where $s_v^{t}$ represents the blur radius (scale) at the $t$-th step. We model feature diffusion as a Markov chain where each state is progressively corrupted:
\begin{equation}
F_v^{t}=\mathcal{B}\!\left(F_v^{t-1}, s_v^{t}\right),\quad t\in\{1,\ldots,T\},
\end{equation}
where $T$ is the total number of diffusion steps. The Gaussian kernel is parameterized by $s_v^{t}$:
\begin{equation}
\mathcal{B}(s_v^{t})=\frac{1}{2\pi(s_v^{t})^2}\exp\!\left(-\frac{i^2+j^2}{2(s_v^{t})^2}\right).
\end{equation}
Intuitively, a larger $s_v^{t}$ induces stronger blurring, attenuating high-frequency details to emulate severe domain degradations.

To adapt the corruption strength to local feature statistics, we introduce a dynamic blur schedule. The base intensity $s_v^{t}$ is adjusted to $(s_v^{t})^{d}$ based on the local mean $\mu_v^{t-1}$ and variance $(\sigma_v^{t-1})^2$ of the feature map:
\begin{equation}
(s_v^{t})^{d}=s_v^{t}\cdot\left(1+\alpha\cdot g\!\left(\mu_v^{t-1},(\sigma_v^{t-1})^2\right)\right),
\end{equation}
where $\alpha$ is set to 0.5 to control the adjustment magnitude, and $g(\cdot)$ is an MLP-based modulation head. The forward diffusion is thus performed as $F_v^{t}=\mathcal{B}(F_v^{t-1}, (s_v^{t})^{d})$.

\noindent\textbf{Text-Guided Reverse Diffusion.}
Instead of blindly removing the blur, we employ a text-guided restoration process to generate ‘restored-yet-shifted’ features. We utilize a learnable deblurring network $\mathcal{D}(\cdot)$ (e.g., a U-Net) that is conditioned on the refined textual guidance $F_t^3$ (from Sec.~\ref{sec:text_chain}). The reverse process iteratively restores the feature:
\begin{equation}
\hat{F}_v^{t-1}=\mathcal{D}\!\left(\hat{F}_v^{t}, (s_v^{t})^{d}, F_t^3, \theta\right),\quad t=T,\ldots,1,
\end{equation}
with initialization $\hat{F}_v^{T}=F_v^{T}$. By injecting $F_t^3$, the network learns to hallucinate semantic details consistent with the text description during the deblurring process.

To ensure the consistency of the restored features with the target style, we enforce a cross-modal diffusion loss on the final output $\hat{F}_v^{0}$:
\begin{equation}
\mathcal{L}_{diff} = 1 - \mathrm{sim}\!\left(\psi(\mathrm{Pool}(\hat{F}_v^{0})),\,F_t^{3}\right).
\end{equation}
This constraint ensures that the generated off-manifold samples preserve the underlying content structure while incorporating the stylistic shifts specified by the text.

\noindent\textbf{Dual-Chain Integration.}
Finally, to synthesize the comprehensive off-manifold sample, we integrate the outputs from both the textual and visual chains. We fuse the diffusion-distorted content $\hat{F}_v^{0}$ with the text-evolved style $\tilde{F}_s$ (from Eq. (8)) and the original content $F_{v}^{0}$:
\begin{equation}
F_e=\mathrm{Conv}\!\left(\mathrm{Concat}(\tilde{F}_s, \hat{F}_v^{0}, F_{v}^{0})\right).
\end{equation}
The resulting $F_e$ encompasses both global style shifts and local structural distortions, serving as a robust hard example for the subsequent Manifold Regression module.

\subsection{Manifold Regression via Prototype Anchoring}
\label{sec3.3}
\noindent\textbf{Geometric Definition of Semantic Manifold.}
Before introducing the prototype anchoring mechanism, we first provide a more explicit definition of the semantic manifold used in this work. 
Given the source-domain ROI-level instance feature set 
$\mathcal{Z}_s=\{(z_i,y_i)\}_{i=1}^{N}$, where $z_i\in\mathbb{R}^{d}$ denotes the instance feature and $y_i$ is its category label, we define the source semantic manifold in the detector feature space as the union of class-conditional semantic supports:
\begin{equation}
\mathcal{M}_s=\bigcup_{c=1}^{C}\mathcal{M}_c,
\end{equation}
where $\mathcal{M}_c$ denotes the semantic support region of category $c$.

Instead of treating a discrete prototype as the complete manifold, we use each class-specific prototype as a local semantic anchor and approximate the corresponding class-conditional support by a prototype-centered neighborhood:
\begin{equation}
\mathcal{M}_c \approx \mathcal{N}(\mathcal{P}_c,r_c)
=
\{z \mid d(\mathcal{H}(z),\mathcal{P}_c)\le r_c\},
\end{equation}
where $\mathcal{P}_c$ is the class-specific prototype of category $c$, and $\mathcal{H}(\cdot)$ denotes the shared ROI-level mapping head that maps instance features into the prototype space. 
The distance function $d(\cdot,\cdot)$ is computed in the normalized prototype space and is implemented as cosine distance in our experiments.

The radius $r_c$ denotes the class-specific semantic tolerance radius, which characterizes the allowable semantic neighborhood around $\mathcal{P}_c$ and reflects normal intra-class variations caused by scale, pose, occlusion, viewpoint, and subcategory differences. 
Specifically, $r_c$ is estimated as a high-percentile statistic of the distances between source-domain ROI features of category $c$ and their corresponding prototype:
\begin{equation}
r_c=
\mathrm{Quantile}_{q}
\left(
\{d(\mathcal{H}(z_i),\mathcal{P}_c)\mid y_i=c\}
\right),
\end{equation}
where $\mathrm{Quantile}_{q}(\cdot)$ denotes the $q$-th percentile, and $q$ is set to $0.95$ by default to cover most normal source-domain intra-class variations. 
This radius defines the geometric neighborhood of each class-specific prototype, rather than introducing an additional learnable parameter or hard training constraint.

Under this definition, a prototype is not regarded as the semantic manifold itself, but serves as a local anchor whose neighborhood approximates the corresponding class-conditional semantic support. In this way, Dual-CoT generates off-manifold deviations to simulate unseen domain shifts, while Manifold Regression learns a parameterized rectification mapping that pulls these deviant features back toward their class-specific prototype neighborhoods. Since the neighborhood constraint is relaxed into the prototype-anchored regression objective, prototype anchoring does not force all instances of the same category to collapse into a single point. Instead, it provides a stable class-level semantic reference while preserving category semantics, instance-level structure, and normal intra-class variations.

\noindent\textbf{Class-Specific Prototype Clustering.}
To construct stable semantic anchors on the source domain manifold, we propose a novel Class-Specific Prototype. 
The motivation for using class-specific prototypes as manifold anchors is that source-domain ROI features of the same category usually concentrate around high-density semantic regions in the normalized detector feature space. 
Thus, each prototype serves as an empirical semantic center of its class-conditional support, and its neighborhood provides a local approximation to the corresponding semantic manifold.
By leveraging soft assignments and residual aggregation, we aggregate features into $K$ learnable cluster centers (where $K$ corresponds to the number of classes), thereby generating robust semantic anchors.

Given the input content feature map $F_c \in \mathbb{R}^{H \times W \times C}$ (derived from Eq. (1)), we first compute the soft assignment scores $\theta \in \mathbb{R}^{H \times W \times K}$, which represent the probability of each spatial location belonging to a specific cluster. This is achieved via a $1\times 1$ convolution followed by a Softmax function:
\begin{equation}\theta = \text{Softmax}(\text{Conv}(F_c)).\end{equation}
With the pixel-wise assignments $\theta$ established, we employ a set of learnable cluster centers $S=\{S_k\}_{k=1}^{K}$ ($S\in\mathbb{R}^{K\times C}$) to serve as semantic reference points. We then aggregate the residuals for each prototype by weighing the feature differences against these centers:
\begin{equation}
V_k = \sum_{x \in \Omega} \theta_{xk}\big(F_{c,x} - S_k\big), \quad k=1,\ldots,K,
\end{equation}
where $\Omega$ denotes the spatial grid and $F_{c,x}\in\mathbb{R}^{C}$ represents the feature at location $x$. 

Subsequently, to obtain compact class-specific prototypes $\mathcal{P} = \{\mathcal{P}_k\}_{k=1}^K$, we apply $\ell_2$ normalization to the aggregated residuals $V_k$, followed by a learnable linear projection $W_p$:
\begin{equation}\mathcal{P}_k = W_p \cdot \frac{V_k}{\|V_k\|_2},\end{equation}
where $W_p \in \mathbb{R}^{C \times C}$ 
denotes the projection weights. These prototypes $\mathcal{P}$ serve as stable semantic anchors for the subsequent regression task.

\noindent\textbf{Prototype-Guided Feature Enhancement.}
To utilize the learned global semantics for detection, we broadcast the prototypes $\mathcal{P}$ back to the spatial dimensions based on the predicted assignment $\theta$ and fuse them with the original input $F_{in}$. A fusion convolution is applied to generate the prototype-enhanced feature map $F_p$:
\begin{equation}
F_p = \text{Conv}(\text{Concat}(F_{in}, \theta \cdot \mathcal{P})).
\end{equation}
This enhanced map $F_p$ suppresses background noise and highlights class-discriminative regions, serving as the refined input for the final object detection heads during the inference.

\noindent\textbf{Manifold Rectification Mechanism.}
Having established the stable prototypes $\mathcal{P}$, we treat them as semantic anchors of the source manifold in the detector feature space. This stage imposes a prototype-guided rectification objective on the Dual-CoT-generated off-manifold features $F_e$ through $\mathcal{H}(\cdot)$.
In implementation, $\mathcal{H}(\cdot)$ is instantiated as a lightweight ROI-level mapping head shared by clean source features and off-manifold features.
Given an ROI feature $z\in\mathbb{R}^{d}$, $\mathcal{H}(\cdot)$ maps it into a $d_p$-dimensional prototype space through a two-layer MLP with ReLU activation, followed by $\ell_2$ normalization.
We set $d_p=256$ in all experiments.
Thus, the mapped instance features $\mathcal{H}(z)$ and the class-specific prototypes $\mathcal{P}_c$ lie in the same normalized prototype space, where cosine distance is used for prototype anchoring and regression.

To ensure precise supervision, we operate at the instance level: using ground-truth boxes, we extract paired features $z_{src}=\mathrm{ROI}(F_{in})$ and $z_{off}=\mathrm{ROI}(F_e)$, and map both into the prototype space via a shared head $\mathcal{H}$. 
From a geometric viewpoint, we define the source semantic manifold as the union of class-conditional prototype neighborhoods 
$\bigcup_{k=1}^{K}\mathcal{N}(\mathcal{P}_k,r_k)$, 
and treat the regression as learning a parametric rectification mapping 
$\mathcal{H}: z_{off} \rightarrow \mathcal{N}(\mathcal{P}_{gt},r_{gt})$ 
under paired supervision.
By doing so, the operator $\mathcal{H}(\cdot)$ learns to collapse shift-induced nuisance factors (e.g., style/weather) orthogonal to the manifold and restore manifold-consistent representations. 
Specifically, the regression loss $\mathcal{L}_{reg}$ in Eq.~(25) serves as an explicit objective for this rectification by pulling $\mathcal{H}(z_{off})$ toward its prototype neighborhood and its paired clean reference $\mathcal{H}(z_{src})$, thereby minimizing the rectification error.

We propose a Hierarchical Manifold Regression objective, which imposes constraints at both global and local geometric levels. To enforce class-discriminative power, the rectified feature $\mathcal{H}(z_{off})$ must align with its corresponding global class prototype $\mathcal{P}_{gt}$. We employ a contrastive formulation to pull the feature towards $\mathcal{P}_{gt}$ while pushing it away from other prototypes $\mathcal{P}_{k \neq gt}$, thereby sharpening the manifold boundaries:
\begin{equation}
\mathcal{L}_{global}
= -\log \frac{
\exp\left(\mathrm{sim}\!\left(\mathcal{H}(z_{off}),\, \mathcal{P}_{gt}\right)/\tau\right)
}{
\sum_{k=1}^{K}
\exp\left(\mathrm{sim}\!\left(\mathcal{H}(z_{off}),\, \mathcal{P}_{k}\right)/\tau\right)
},
\end{equation}
where $\tau$ is a temperature hyperparameter controlling the concentration of the manifold, empirically set to 0.1.

While global anchoring ensures correct classification, it risks discarding instance-specific structural details. To fulfill the objective of error correction, we explicitly minimize the residual between the mapped off-manifold feature and its clean source counterpart. This explicitly counteracts the domain shifts injected by Dual-CoT:
\begin{equation}
\mathcal{L}_{local} = \| \mathcal{H}(z_{off}) - \text{sg}(\mathcal{H}(z_{src})) \|_2^2,
\end{equation}
where $\| \cdot \|_2^2$ denotes the squared $\ell_2$ norm, and $\text{sg}(\cdot)$ represents the stop-gradient operation, ensuring that the clean source features serve as stable regression targets. The final Manifold Regression loss is defined as a weighted sum:
\begin{equation} \mathcal{L}_{reg} = \mathcal{L}_{global} + \lambda_{loc} \mathcal{L}_{local}, \end{equation} 
where $\lambda_{loc}$ is a trade-off hyperparameter that balances global semantic separability with local structural preservation, empirically set to 0.5. By minimizing $\mathcal{L}_{reg}$, the model learns a directional correction mapping $\mathcal{H}(\cdot)$ that rectifies perturbed off-manifold features back to manifold-consistent neighborhoods, rather than merely promoting prototype proximity as in standard metric/prototype alignment.

\noindent\textbf{Total Optimization Objective.}
To synergistically integrate feature disentanglement, off-manifold generation, and manifold rectification, we optimize MR-DCoT under a unified multi-task learning objective. To avoid redundant top-level hyper-parameter tuning, we simply sum the loss terms without introducing extra global weighting coefficients, since the required trade-offs are already handled by the module-specific factors. The overall objective is formulated as:
\begin{equation}
\mathcal{L}_{total} = \mathcal{L}_{det} + \mathcal{L}_{emb} + \mathcal{L}_{align} + \mathcal{L}_{diff} + \mathcal{L}_{reg},
\end{equation}
where $\mathcal{L}_{det}$ denotes the standard detection loss. The auxiliary terms include the disentangled embedding loss $\mathcal{L}_{emb}$ (Eq.~(4)), the style alignment loss $\mathcal{L}_{align}$ (Eq.~(9)), the cross-modal diffusion consistency loss $\mathcal{L}_{diff}$ (Eq.~(14)), and the manifold rectification loss $\mathcal{L}_{reg}$ (Eq.~(25)). This formulation encourages the network to jointly learn disentangled representations, synthesize structured off-manifold outliers, and acquire a robust error-correction mechanism that rectifies deviant features back to the class-semantic manifold.

\noindent\textbf{Training and Inference.}
During the training phase, all components are jointly optimized in an end-to-end manner. Specifically, the Dual-CoT module is activated to generate off-manifold features $F_e$, which serve as structured proxies to guide the learning of the rectification operator. To guarantee robust generalization, the detection head is optimized on a dual-stream input, simultaneously processing both the prototype-enhanced source features $F_p$ and the off-manifold features $F_e$. Concurrently, the class-specific prototypes $\mathcal{P}$ are instantiated as stable manifold anchors within the regression objective. In contrast, during the inference phase, the Dual-CoT module and all auxiliary objectives are deactivated. The detector performs a standard single forward pass, leveraging the robust representations learned during training, specifically the prototype-enhanced features $F_p$, to produce predictions.

\subsection{Further Discussion}
\label{sec:discussion}

In this section, we further discuss our manifold regression paradigm and make a comparison with our preliminary conference version, SE-COT~\cite{SECOT}.

\textbf{Necessity of Manifold Regression.} Recent advancements in image generation by Li et al.~\cite{back} demonstrate that natural images in high-dimensional observation spaces are not randomly distributed but concentrate near low-dimensional manifolds. Inspired by this, we posit that source-domain representations admit a compact, low-dimensional semantic support in the detector feature space, whereas unknown domain shifts (e.g., transitioning from sunny to rainy conditions) perturb samples into expansive off-manifold regions. Accordingly, we explicitly learn a parametric rectification operator that regresses off-manifold features back to manifold-consistent semantic coordinates.
Building upon this view, we argue that rather than exhaustively covering infinite variations via finite simulations, a more robust strategy is to learn an error-correction mapping. Concretely, we instantiate the semantic manifold using class-specific prototypes as local anchors and formulate generalization as a class-conditional manifold regression problem. Given an off-manifold feature, the regressor is trained to (i) regress it toward its corresponding prototype neighborhood and (ii) minimize its residual to the paired clean-source feature, thereby learning a directional correction rule instead of a static alignment. This design ensures that even for unforeseen shifts that were never explicitly simulated, the detector can actively rectify deviant representations back into a safe semantic subspace, moving beyond perturbation enumeration toward a rectification-based correction paradigm.

\textbf{Synergy of Dual-CoT and Manifold Anchoring.} The innovation of MR-DCoT lies not merely in the regression mechanism itself, but in how we construct the off-manifold samples to train this regressor. A robust rectification operator necessitates training on hard, structured, and semantically consistent outliers. Relying solely on a single modality is inadequate: while text provides high-level semantic guidance (e.g., ‘rainy’), it lacks structural granularity; conversely, simple visual noise suffers from a lack of semantic controllability. Our Dual-CoT bridges this gap. The Textual Chain defines the high-level evolution direction, while the Visual Chain injects fine-grained structural perturbations via diffusion. This synergy generates structured off-manifold proxies, where samples exhibit deviations that effectively simulate real-world domain shifts, yet their semantics remain sufficiently preserved to enable prototype anchoring. This provides the Manifold Regression module with sufficiently realistic simulation samples, ensuring that the learned rectification operator can handle both stylistic variations and structural corruptions.

\textbf{Comparison with SE-COT.}
This work is a substantial extension of our conference version, SE-COT~\cite{SECOT}.
To make the relationship clearer, SE-COT can be regarded as the textual-chain-based simulation foundation of this work, while MR-DCoT further extends it toward a visual-text off-manifold simulation and manifold rectification framework.
MR-DCoT differs significantly from SE-COT in three key aspects:
(1) \textbf{Paradigm Shift:} SE-COT follows a Simulation-Driven paradigm, focusing primarily on expanding semantic diversity via VLM-generated prompts to cover more domains. In contrast, MR-DCoT introduces a Simulation-Regression closed loop. We move beyond mere expansion to explicitly modeling the rectification of deviant features, reformulating generalization as a manifold regression problem.
Thus, the goal is no longer only to generate more domain styles, but to learn how to correct features once they deviate from the source semantic support.
(2) \textbf{Generation Mechanism:} SE-COT relies solely on textual prompts to guide style generation, which may overlook local structural variations (e.g., blurs, noise). MR-DCoT upgrades this to a Visual-Text Dual Chain, integrating diffusion-based visual perturbations with textual evolution. This allows us to simulate more complex, coupled domain shifts (e.g., stormy weather with visibility degradation) that SE-COT cannot handle.
In other words, the Textual Chain inherits the semantic evolution ability of SE-COT, while the newly introduced Visual Chain complements it with fine-grained structural degradation.
(3) \textbf{Optimization Objective:} SE-COT optimizes for semantic alignment in the expanded space. MR-DCoT introduces the Class-Specific Prototype Anchoring mechanism, which imposes a prototype-anchored rectification objective to guide off-manifold features toward their corresponding class-conditional prototype neighborhoods, resulting in superior stability and discriminability.
Therefore, MR-DCoT differs from SE-COT not only in how off-manifold samples are generated, but also in how these samples are used: they serve as structured supervision for learning a prototype-anchored rectification mapping.

\vspace{-5pt}
\section{Experiments}
To comprehensively evaluate the effectiveness of the proposed method, we conduct extensive experiments on two distinct tasks: Single-Domain Generalized Object Detection and Zero-shot Domain Adaptation for semantic segmentation. For Single-DGOD, we follow the established settings in \cite{S_DG, pdoc, UFR} to validate the model's generalization capability. Furthermore, we evaluate our method on the Reality-to-Art benchmark \cite{Div}. To further demonstrate the versatility of our approach, we incorporate Zero-shot Domain Adaptation (ZSDA) experiments, following the protocol of P\O DA \cite{Poda}, as an auxiliary validation for the semantic segmentation task.
\vspace{-10pt}
\subsection{Experimental setup}
\textbf{Dataset. }
Our experiments are conducted on three complementary benchmarks, covering (i) single-domain generalized object detection under adverse driving conditions, (ii) generalization from reality to art, and (iii) zero-shot domain adaptation for semantic segmentation.

\textbf{Diverse Driving Weather Scenarios.}
To ensure fair comparisons with prior Single-Domain Generalized Object Detection (Single-DGOD) studies, we follow the same dataset protocol as UFR \cite{UFR} and DIV \cite{Div}. The benchmark contains five subsets, each corresponding to a distinct weather condition: Day Clear, Night Sunny, Dusk Rainy, Night Rainy, and Day Foggy. In our setup, the detector is trained exclusively on Day Clear, which provides a stable source-domain distribution with clear illumination and diverse traffic-related objects (e.g., vehicles and pedestrians). After training, the model is directly evaluated on the remaining four subsets to measure generalization under domain shifts induced by illumination changes (Night Sunny), precipitation (Dusk Rainy and Night Rainy), and reduced visibility (Day Foggy).

\textbf{Generalization from Reality to Art.}
Following \cite{Art, Div}, we use the Pascal VOC2007 and VOC2012 trainval sets \cite{VOC} for training, totaling 16,551 images. For model selection, we validate on the VOC2007 test set (5,000 images) and choose the best-performing checkpoint on the source domain. The selected model is then evaluated on three unseen artistic target domains: Clipart1k, Watercolor2k, and Comic2k. The Clipart dataset contains the same 20 categories as PASCAL VOC, while Watercolor2k and Comic2k each include 6 classes that form subsets of the VOC categories.

\textbf{Zero-shot Domain Adaptation (ZSDA) for Semantic Segmentation.}
Following PØDA \cite{Poda}, we use Cityscapes \cite{Cityscapes} for training, totaling 2,975 images annotated with 19 categories. The model is evaluated on two unseen target domains: ACDC \cite{ACDC} and GTA5 \cite{GTA5}. ACDC contains urban scenes captured under adverse conditions, while GTA5 serves as a benchmark for real-to-synthetic adaptation. Additionally, we evaluate the synthetic-to-real setting by training on GTA5 and testing on Cityscapes. All evaluations are performed on the official validation sets, except for GTA5, where we use a random subset of 1,000 images.

\textbf{Metric.}
To ensure fair comparisons with prior work, we follow the standard evaluation protocols used in Single-DGOD \cite{S_DG, UFR} for object detection, and the common practice in zero-shot domain adaptation for semantic segmentation.
Object detection (Single-DGOD):
We adopt the same evaluation metrics as existing Single-DGOD studies \cite{S_DG, UFR}. Detection performance is quantified using mean Average Precision (mAP) at an IoU threshold of 0.5, computed over all categories on each dataset/target domain. 
Semantic segmentation (ZSDA):
For the zero-shot domain adaptation (ZSDA) semantic segmentation experiments, we evaluate adaptation quality using the mean Intersection-over-Union (mIoU, \%), computed over the standard semantic classes. All models are evaluated on target-domain images at their original resolutions.
\begin{table}[!t]
  \centering
  \caption{Single-Domain Generalization Results (mAP(\%)) in Diverse Weather Scenarios. We evaluate our method across different backbones and frameworks. \textbf{The bold} represent the best results.}
  \fontsize{18}{10}\selectfont 
  \setlength{\tabcolsep}{1.5pt}
  \resizebox{1.0\linewidth}{!}{
    \begin{tabular}{l|c|c|cccc}
    \toprule
    \toprule
    \multirow{2}[4]{*}{\textbf{Method}} & \multirow{2}[4]{*}{\textbf{Backbone}} & \multicolumn{1}{c|}{\textbf{Source}} & \multicolumn{4}{c}{\textbf{Unseen Target Domains}} \\
    \cmidrule{3-7}          &       & Day Clear & Night Sunny & Dusk Rainy & Night Rainy & Day Foggy \\
    \midrule
    \multicolumn{7}{c}{\textit{\textbf{Faster R-CNN Framework}}} \\
    Faster R-CNN \cite{ren2015faster} & R-101 & 48.1  & 34.4  & 26.0  & 12.4  & 32.0  \\
    SW \cite{SW} & R-101 & 50.6  & 33.4  & 26.3  & 13.7  & 30.8  \\
    ISW \cite{ISW} & R-101 & 51.3  & 33.2  & 25.9  & 14.1  & 31.8  \\
    S-DGOD \cite{S_DG} & R-101 & 56.1  & 36.6  & 28.2  & 16.6  & 33.5  \\
    C-Gap \cite{C_Cap} & R-101 & 51.3  & 36.9  & 32.3  & 18.7  & 38.5  \\
    PDOC \cite{pdoc} & R-101 & 53.6  & 38.5  & 33.7  & 19.2  & 39.1  \\
    UFR \cite{UFR} & R-101 & 58.6  & 40.8  & 33.2  & 19.2  & 39.6  \\
    DIV \cite{Div} & R-101 & 52.8  & 42.5  & 38.1  & 24.1  & 37.2  \\
    G-NAS \cite{G-NAS} & R-101 & 58.4  & 45.0  & 35.1  & 17.4  & 36.4  \\
    SRCD \cite{srcd} & R-101 & --    & 36.7  & 28.8  & 17.0  & 35.9  \\
    FWCL \cite{FWCL} & R-101 & 55.5  & 37.5  & 32.6  & 18.9  & 32.3  \\
    SE-COT \cite{SECOT} & R-101 & 55.4  & 42.0  & 39.2  & 24.5  & 40.6  \\
    \textbf{Ours} & R-101 & 56.2  & 47.4  & 42.2  & 28.1  & 42.8  \\
    SE-COT \cite{SECOT} & Swin-T & 64.4 & 52.7 & 49.5 & 33.7 & 44.9 \\
    \textbf{Ours} & Swin-T & \textbf{65.1}  & \textbf{54.6}  & \textbf{53.2}  & \textbf{36.4}  & \textbf{48.3}  \\
    \midrule

    \multicolumn{7}{c}{\textit{\textbf{YOLO Framework}}} \\
    YOLOv10-L \cite{yolov10} & CSPNet & 60.8  & 43.8  & 35.4  & 21.2  & 38.7  \\
    SE-COT \cite{SECOT} & CSPNet & 60.4 & 46.2 & 39.5 & 23.9 & 41.4 \\
    \textbf{Ours} & CSPNet & \textbf{63.7}  & \textbf{50.3}  & \textbf{42.5}  & \textbf{27.0}  & \textbf{43.9}  \\
    \midrule

    \multicolumn{7}{c}{\textit{\textbf{Generative Framework}}}  \\
    DiffusionDet \cite{diffusiondet} & Swin-T & 65.2  & 50.8  & 47.6  & 27.3  & 42.7  \\
    SE-COT \cite{SECOT} & Swin-T & 67.4 & 54.1 & 49.7 & 31.2 & 46.6 \\
     \textbf{Ours} & Swin-T &\textbf{68.3}  & \textbf{55.4} & \textbf{52.1} & \textbf{33.2} & \textbf{48.7}    \\   
    \midrule
    \multicolumn{7}{c}{\textit{\textbf{Vision-Language Framework}}} \\
    GLIP-T \cite{GLIP} & Swin-T & 61.2 & 44.3 & 42.5 & 25.6 & 39.7  \\
    PGST \cite{PGST}   & Swin-T & 63.7  & 47.9  & 44.5  & 28.4  & 42.5  \\
    SE-COT \cite{SECOT} & Swin-T & 63.1 & 48.5 & 43.9 & 29.1 & 41.7 \\
    \textbf{Ours} & Swin-T & \textbf{65.3} & \textbf{50.2} & \textbf{45.5} & \textbf{29.7} & \textbf{46.9} \\
    \midrule

    \multicolumn{7}{c}{\textit{\textbf{DETR Framework}}} \\
    DINO \cite{DINO} & DINOv2 & 65.3 & 52.5 & 51.4 & 41.7 & 55.2 \\
    Cauvis \cite{Cauvis} & DINOv2 & \textbf{73.7} & 61.2 & \textbf{64.6} & \textbf{47.6} & 56.5 \\
    SE-COT \cite{SECOT} & DINOv2 & 69.9 & 59.2 & 60.1 & 43.8 & 57.2 \\
    \textbf{Ours} & DINOv2 & 71.2 & \textbf{62.4} & 63.7 & 46.9 & \textbf{58.2} \\
    \bottomrule
    \bottomrule
    \end{tabular}%
  }
  \vspace{-10pt}
  \label{tab1}%
\end{table}%

\subsection{Implementation Details}
To evaluate the effectiveness of our method, we conducted fair comparisons across a diverse set of baseline models, including Faster R-CNN \cite{ren2015faster}, YOLO \cite{yolov10}, DiffusionDet \cite{diffusiondet}, GLIP\cite{GLIP}, and DINO \cite{DINO}. Additionally, we report detection results using multiple backbone architectures.
For Disentangled Feature Embedding, we set the style consistency weight $\lambda_{sc} = 0.5$ to balance orthogonality with source priors. In the Visual-Text Dual-CoT module, we utilize a pre-trained CLIP \cite{CLIP} as the text encoder $E_{text}$; the detector is initialized with frozen CLIP weights to preserve semantic alignment, while the visual chain employs a dynamic blur magnitude of $\alpha = 0.5$  for effective structural perturbation. For Manifold Regression, we set the temperature $\tau = 0.1$ to ensure manifold compactness and the local regression weight $\lambda_{loc} = 0.5$  to balance structural rectification with semantic separability.
Model optimization is performed using stochastic gradient descent (SGD) with a learning rate of 1.0, momentum of 0.9, and weight decay of 0.0005. All experiments are conducted on four NVIDIA RTX 4090 GPUs with a batch size of 8.

\begin{figure*}[t]
  \centering
  \vspace{-0.3in}
  \includegraphics[width=2.0\columnwidth]{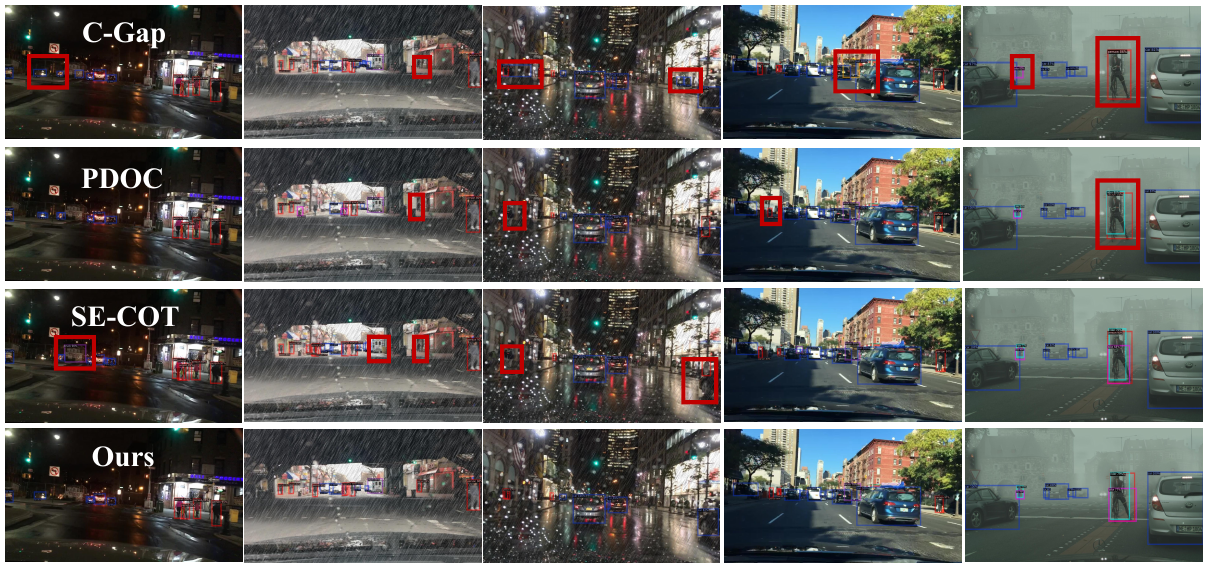}
  \caption{\textbf{Qualitative Results}: Detection results under different weather conditions. The four rows display the results from C-Gap \cite{C_Cap}, PDOC \cite{pdoc}, SE-COT \cite{SECOT}, and our method, respectively. Objects highlighted in \textbf{red boxes} indicate missed detections or misclassifications.}
    \label{F5}
    \vspace{-0.2in}
\end{figure*}

\begin{table}[!t]
  \centering
  \caption{Per-class results (\%) on Day Clear to Dusk Rainy. \textbf{The bold} represents the best results.}
   \vspace{-5pt}
  \fontsize{8}{6}\selectfont
  \resizebox{\linewidth}{!}{
    \begin{tabular}{cc|ccccccc|c}
    \toprule
    \toprule
    \multicolumn{2}{c|}{\multirow{2}[4]{*}{Method}} & \multicolumn{7}{c|}{AP}                      & \multicolumn{1}{c}{mAP} \\
\cmidrule{3-10}    \multicolumn{2}{c|}{} & Bus & Bike & Car & Motor & Person & Rider & Truck & All \\
    \midrule
    \multicolumn{2}{c|}{Faster R-CNN \cite{ren2015faster}} & 28.50  & 20.30  & 58.20  & 6.50  & 23.40  & 11.30  & 33.90  & 26.00  \\
    \multicolumn{2}{c|}{S-DGOD \cite{S_DG}} & 37.10  & 19.60  & 50.90  & 13.40  & 19.70  & 16.30  & 40.70  & 28.20  \\
    \multicolumn{2}{c|}{C-Gap \cite{C_Cap}} & 37.80  & 22.80  & 60.70  & 16.80  & 26.80  & 18.70  & 42.40  & 32.30  \\ 
    \multicolumn{2}{c|}{PDOC \cite{pdoc}} & 39.40  & 25.20  & 60.90  & 20.40  & 29.90  & 16.50  & 43.90  & 33.70  \\
    \multicolumn{2}{c|}{UFR \cite{UFR}} & 37.10  & 21.80  & 67.90 & 16.40  & 27.40  & 17.90  & 43.90  & 33.20  \\
    \multicolumn{2}{c|}{YOLOv10-L \cite{yolov10}} & 41.30 & 21.30 & 69.90 & 16.50 & 32.80 & 18.40 & 47.90  & 35.40  \\
    \multicolumn{2}{c|}{FWCL\cite{FWCL}} & 37.20 & 22.60 & 60.50 & 17.00 & 27.00 & 19.90 & 43.80 & 32.60\\
    \multicolumn{2}{c|}{SE-COT\cite{SECOT}(Res101)} & 45.75 & 19.20 & 74.32 & 23.43 & 35.81 & 22.14 & 53.47 & 39.16 \\
    \multicolumn{2}{c|}{SE-COT\cite{SECOT}(Swin-T)} & 57.34 & 40.23 & 76.76 & 28.36 & 50.03 & 32.66 & 60.80 & 49.45 \\
    \midrule
    \multicolumn{2}{c|}{Ours(Res101)} & 53.27 & 27.63 & 59.32 & 24.98 & 43.10 & 27.31 & 59.63 & 42.18 \\  
    \multicolumn{2}{c|}{Ours(Swin-T)} & \textbf{61.57} & \textbf{41.04} & \textbf{79.05} & \textbf{31.30} & \textbf{57.31} & \textbf{37.40} & \textbf{64.82} & \textbf{53.21} \\  
    \bottomrule
    \bottomrule
    \end{tabular}%
    }
  \label{tab3}
  \vspace{-5pt}
\end{table}

\subsection{Comparison with the State of the Art}
\textbf{Performance on Diverse Weather Scenarios.} We present a comprehensive comparative evaluation against state-of-the-art Single-DGOD approaches in Table \ref{tab1}. Our method achieves superior detection performance across a diverse range of baselines, including Faster R-CNN \cite{ren2015faster}, YOLO \cite{yolov10}, DiffusionDet \cite{diffusiondet}, GLIP \cite{GLIP}, and DINO \cite{DINO}, effectively validating its effectiveness and universality. To facilitate fair comparisons, we further report results utilizing multiple backbone architectures. Figure \ref{F5} provides a qualitative visual comparison using the Faster R-CNN with a ResNet-101, benchmarking against three representative competitors: C-Gap \cite{C_Cap}, PDOC \cite{pdoc}, and SE-COT \cite{SECOT}. As illustrated, while competing methods suffer from frequent missed detections and misclassifications in challenging low-visibility conditions such as rainy nights, our model maintains robust detection capabilities, accurately localizing objects that other methods fail to recall.

\textbf{Generalization across Diverse Detection Frameworks.} To rigorously validate the universality and architectural agnosticism of our approach, we extend our evaluation to encompass a broad spectrum of modern detection paradigms, as summarized in Table \ref{tab1}. Specifically, we integrate our method into five representative architectures: (1) Faster R-CNN \cite{ren2015faster}, representing classic two-stage detectors; (2) YOLOv10-L \cite{yolov10}, representing state-of-the-art one-stage real-time detectors; (3) DiffusionDet \cite{diffusiondet}, representing the emerging generative detection paradigm; (4) GLIP-T \cite{GLIP}, representing vision-language pre-training frameworks; and (5) DINO \cite{DINO}, representing end-to-end detection transformers. Experimental results demonstrate that our method consistently boosts detection performance across CNN-based, Transformer-based, and diffusion-based architectures, effectively validating its capability to enhance generalization across multiple benchmarks.

\begin{table}[t]
  \centering
  \caption{Per-class results (\%) on Day Clear to Night Sunny. \textbf{The bold} represents the best results.}
    \vspace{-5pt}
  \fontsize{8}{6}\selectfont
  \resizebox{\linewidth}{!}{
    \begin{tabular}{cc|ccccccc|c}
    \toprule
    \toprule
      \multicolumn{2}{c|}{\multirow{2}[4]{*}{Method}} & \multicolumn{7}{c|}{AP} & \multicolumn{1}{c}{mAP} \\
\cmidrule{3-10}    \multicolumn{2}{c|}{} & Bus & Bike & Car & Motor & Person & Rider & Truck & All \\
    \midrule
    \multicolumn{2}{c|}{Faster R-CNN \cite{ren2015faster}} & 34.70  & 32.00  & 56.60  & 13.60  & 37.40  & 27.60  & 38.60  & 34.40  \\
    \multicolumn{2}{c|}{S-DGOD \cite{S_DG}} & 40.60  & 35.10  & 50.70  & 19.70  & 34.70  & 32.10  & 43.40  & 36.60  \\
    \multicolumn{2}{c|}{C-Gap \cite{C_Cap}} & 37.70  & 34.30  & 58.00  & 19.20  & 37.60  & 28.50  & 42.90  & 36.90  \\
    \multicolumn{2}{c|}{PDOC \cite{pdoc}} & 40.90  & 35.00  & 59.00  & 21.30  & 40.40  & 29.90  & 42.90  & 38.50  \\
    \multicolumn{2}{c|}{UFR \cite{UFR}} & 43.60  & 38.10  & 66.10 & 14.70  & 49.10  & 26.40  & 47.50  & 40.80  \\
    \multicolumn{2}{c|}{YOLOv10-L \cite{yolov10}} & 45.30 & 39.10 & 68.90 & 22.40 & 53.80 & 27.30 & 49.80  & 43.80  \\
    \multicolumn{2}{c|}{FWCL\cite{FWCL}} & 39.60 & 36.40 & 58.90 & 18.00 & 42.60 & 26.20 & 40.50 & 37.50 \\
    \multicolumn{2}{c|}{SE-COT(R-101)\cite{SECOT}} & 41.95 & 37.21 & 66.19 & 21.56 & 49.61 & 33.25 & 44.24 & 42.00 \\
    \multicolumn{2}{c|}{SE-COT(Swin-T)\cite{SECOT}}& \textbf{55.54} & 49.66  & 71.00 & 32.87  & 58.75 & \textbf{43.66} & 57.75 & 52.75 \\ 
    \midrule
    \multicolumn{2}{c|}{Ours(Res101)} & 51.30 & 42.06 & 69.30 & 31.12 & 50.35 & 32.13 & 55.42 & 47.38 \\
    \multicolumn{2}{c|}{Ours(Swin-T)} & 50.31 & \textbf{53.24} & \textbf{80.13} & \textbf{33.52} & \textbf{63.21} & 41.20 & \textbf{60.83} & \textbf{54.63} \\
    \bottomrule
    \bottomrule
    \end{tabular}%
    }
  \label{tab4}%
    \vspace{-12pt}
\end{table}

\textbf{Quantitative Performance Analysis.} As detailed in Table \ref{tab1}, our method consistently achieves leading performance across a diverse range of frameworks. Specifically, under the Faster R-CNN framework with identical backbone settings, our method yields a significant improvement of +3.6\% mAP (rising from 24.5\% to 28.1\%) over SE-COT in the challenging Night Rainy scenario, which suffers from severe visibility degradation. Similarly, we observe a +3.1\% mAP gain under the YOLO framework. Furthermore, in terms of architectural adaptability, our approach consistently boosts the performance of advanced vision-language models. On the GLIP-T baseline, our method outperforms the previous state-of-the-art PGST by +2.3\% in the Night Sunny domain (50.2\% vs. 47.9\%) and by +4.4\% in the Day Foggy domain (46.9\% vs. 42.5\%). Finally, when scaled to large foundation models such as DINO \cite{DINO}, our method remains highly competitive, effectively validating its robust adaptability and generalization capabilities across diverse detection frameworks.

\textbf{Day Clear to Dusk Rainy.} The transition to the Dusk Rainy domain introduces compound challenges: diminishing illumination degrades global visibility, while rain streaks cause local visual corruption. Despite these adverse conditions, as presented in Table \ref{tab3}, our method demonstrates exceptional robustness. Under identical backbone settings, it outperforms SE-COT by +3.8\%, effectively validating our method's generalization capability on unseen domains. This success is attributed to our framework design, which not only leverages Dual-CoT to simulate complex domain shifts but also employs manifold regression to acquire robust and invariant features.

\textbf{Day Clear to Night Sunny.} The dramatic shift to the nighttime domain introduces severe challenges, where insufficient illumination and glare interference result in extremely low signal-to-noise ratios and consequent feature degradation. Despite these difficulties, Table \ref{tab4} confirms the superior generalization capability of our model. Specifically, our method achieves a peak mAP of 54.63\%. Notably, in critical categories such as ‘Car’, it attains an AP of 80.13\%, thereby validating the robustness of our approach under conditions of severe illumination deficiency.

\begin{table}[!t]
  \centering
    \vspace{-5pt}
  \caption{Per-class results (\%) on Day Clear to Night Rainy. \textbf{The bold} represents the best results.}
   \vspace{-5pt}
  \fontsize{8}{6}\selectfont
  \resizebox{\linewidth}{!}{
    \begin{tabular}{cc|ccccccc|c}
    \toprule
    \toprule
    \multicolumn{2}{c|}{\multirow{2}[4]{*}{Method}} & \multicolumn{7}{c|}{AP} & \multicolumn{1}{c}{mAP} \\
\cmidrule{3-10}    \multicolumn{2}{c|}{} & Bus & Bike & Car & Motor & Person & Rider & Truck & All \\
    \midrule
    \multicolumn{2}{c|}{Faster R-CNN \cite{ren2015faster}} & 16.80  & 6.90  & 26.30  & 0.60  & 11.60  & 9.40  & 15.40  & 12.40  \\
    \multicolumn{2}{c|}{S-DGOD \cite{S_DG}} & 24.40  & 11.60  & 29.50  & 9.80  & 10.50  & 11.40  & 19.20  & 16.60  \\
    \multicolumn{2}{c|}{C-Gap \cite{C_Cap}} & 28.60  & 12.10  & 36.10  & 9.20  & 12.30  & 9.60  & 22.90  & 18.70  \\
    \multicolumn{2}{c|}{PDOC \cite{pdoc}} & 25.60  & 12.10  & 35.80  & 10.10  & 14.20  & 12.90  & 22.90  & 19.20  \\
    \multicolumn{2}{c|}{UFR \cite{UFR}} & 29.90 & 11.80  & 36.10 & 6.40  & 13.10  & 10.50  & 23.30  & 19.20  \\
    \multicolumn{2}{c|}{YOLOv10-L \cite{yolov10}} & 35.40 & 13.90 & 38.50 & 9.90 & 12.10 & 10.10 & 28.70  & 21.20  \\
    \multicolumn{2}{c|}{FWCL(Res101)\cite{FWCL}} & 28.90 & 11.20 & 34.60 & 9.10 & 12.10 & 12.00 & 23.90 & 18.90 \\
    \multicolumn{2}{c|}{SE-COT(Res101)\cite{SECOT}} & 37.85 & 16.98 & 42.59 & 12.89 & 16.67 & 14.20 & 30.55 & 24.53 \\
    \multicolumn{2}{c|}{SE-COT(Swin-T)\cite{SECOT}} & 49.28 & 19.11 & 56.74 & 11.12 & 31.33 & 20.86 & 47.79  & 33.74 \\
    \midrule
    \multicolumn{2}{c|}{Ours(Res101)} & 40.30 & 19.21 & 35.43 & \textbf{21.40} & 26.89 & \textbf{21.30} & 31.90 & 28.06\\
    \multicolumn{2}{c|}{Ours(Swin-T)} & \textbf{50.30} & \textbf{21.40} & \textbf{60.32} & 17.63 & \textbf{33.29} & 20.10 & \textbf{52.04} & \textbf{36.44} \\
    \bottomrule
    \bottomrule
    \end{tabular}%
    }
     \vspace{-5pt}
  \label{tab5}%
\end{table}

\begin{table}[!h]
  \centering
  \vspace{-5pt}
  \caption{Per-class results (\%) on Day Clear to Day Foggy. \textbf{The bold} represents the best results.}
    \vspace{-5pt}
  \fontsize{8}{6}\selectfont
  \resizebox{\linewidth}{!}{
    \begin{tabular}{cc|ccccccc|c}
    \toprule
    \toprule
      \multicolumn{2}{c|}{\multirow{2}[4]{*}{Method}} & \multicolumn{7}{c|}{AP} & \multicolumn{1}{c}{mAP} \\
\cmidrule{3-10}    \multicolumn{2}{c|}{} & Bus & Bike & Car & Motor & Person & Rider & Truck & All \\
    \midrule
    \multicolumn{2}{c|}{Faster R-CNN \cite{ren2015faster}} & 28.10 & 29.70 & 49.70 & 26.30 & 33.20 & 35.50 & 21.50 & 32.00 \\
    \multicolumn{2}{c|}{S-DGOD \cite{S_DG}} & 32.90 & 28.00 & 48.80 & 29.80 & 32.50 & 38.20 & 24.10 & 33.50 \\
    \multicolumn{2}{c|}{C-Gap \cite{C_Cap}} & 36.10 & 34.30 & 58.00 & 33.10 & 39.00 & 43.90 & 25.10 & 38.50 \\
    \multicolumn{2}{c|}{PDOC \cite{pdoc}} & 36.10 & 34.50 & 58.40 & 33.30 & 40.50 & 44.20 & 26.20 & 39.10 \\
    \multicolumn{2}{c|}{UFR \cite{UFR}} & 36.90 & 35.80 & 61.70 & 33.70 & 39.50 & 42.20 & 27.50 & 39.60 \\
    \multicolumn{2}{c|}{G-NAS \cite{G-NAS}} & 32.40 & 31.20 & 57.70 & 31.90 & 38.60 & 38.50 & 24.50 & 36.40 \\
    \multicolumn{2}{c|}{YOLOv10-L \cite{yolov10}} & 35.70 & 31.30 & 63.90 & 33.30 & 41.10 & 42.60 & 23.40 & 38.70 \\
    \multicolumn{2}{c|}{SE-COT(Res101)\cite{SECOT}} & 39.33 & 36.28 & 60.82 & 33.79 & 39.41 & 42.68 & 31.77 & 40.58 \\
    \multicolumn{2}{c|}{SE-COT(Swin-T)\cite{SECOT}} & 42.11 & 38.32 & 62.79 & \textbf{40.75} & 47.48 & 48.04 & 35.41 & 44.99 \\
    \midrule
    \multicolumn{2}{c|}{Ours(Res101)} & 39.12 & \textbf{45.65} & 63.13 & 31.22 & 42.18 & 40.01 & \textbf{38.47} & 42.83 \\
    \multicolumn{2}{c|}{Ours(Swin-T)} & \textbf{44.31} & 40.35 & \textbf{67.31} & 38.01 & \textbf{50.32} & \textbf{59.13} & 38.45 & \textbf{48.27} \\
    \bottomrule
    \bottomrule
    \end{tabular}%
    } 
  \label{table2}%
\end{table}

\textbf{Day Clear to Night Rainy.} This scenario constitutes the most challenging domain shift, combining severe low-light conditions, heavy rain streak occlusions, and complex surface reflections, which drastically degrade feature discriminability. As shown in Table \ref{tab5}, our method achieves the highest mAP of 36.44\%, surpassing the second-best SE-COT by +2.70\%. Simultaneously, it secures the best performance on critical categories such as ‘Car’ and ‘Person’, validating the effectiveness of our method under extreme domain shifts.

\textbf{Day Clear to Day Foggy.} Foggy weather induces severe light scattering, resulting in a whitening effect and low contrast that drastically blurs object boundaries and obscures distant details. As presented in Table \ref{table2}, our method achieves leading performance across both evaluated backbone architectures. Notably, when utilizing the Swin Transformer, it surpasses SE-COT by +3.28\%. This validates that our MR-DCoT framework not only effectively simulates global hazy and low-contrast shifts but also leverages manifold regression to further enhance the model's generalization capability.
\begin{table}[t]
  \centering
  \caption{Single-Domain Generalization Results (mAP(\%)) from Real to Artistic. \textbf{The bold} represents the best results.}
    \vspace{-5pt}
  \fontsize{8}{4}\selectfont
  \resizebox{\linewidth}{!}{
    \begin{tabular}{cc|c|ccc}
    \toprule
    \toprule
    \multicolumn{2}{c|}{\multirow{2}[4]{*}{Method}} & \multicolumn{1}{c}{} &        & \multicolumn{1}{c}{mAP} &  \\
\cmidrule{3-6}    \multicolumn{2}{c|}{} & \multicolumn{1}{c|}{VOC} & \multicolumn{1}{c}{Comic} & \multicolumn{1}{c}{Watercolor} & \multicolumn{1}{c}{Clipart} \\
    \midrule
    \multicolumn{2}{c|}{Faster R-CNN\cite{ren2015faster}} & 80.4  & 19.4  & 45.6  & 26.5  \\
    \multicolumn{2}{c|}{NP\cite{np}} & 79.2  & 28.9  & 53.3  & 35.4  \\
    \multicolumn{2}{c|}{C-Gap\cite{C_Cap}} & 80.5  & 29.4  & 50.7  & 36.7  \\
    \multicolumn{2}{c|}{DIV\cite{Div}} & 80.1  & 33.2  & 57.4  & 38.9  \\
    \multicolumn{2}{c|}{SE-COT(Res101) \cite{SECOT}} & 82.9  & 34.8  & 57.5  & 40.2  \\
    \multicolumn{2}{c|}{SE-COT(Swin-T) \cite{SECOT}} & 87.6 & 36.9 & 60.7 & 42.5  \\
    \midrule
     \multicolumn{2}{c|}{Ours(Res101)} & 85.3    & 37.3 & 60.3    &43.1  \\
    \multicolumn{2}{c|}{Ours(Swin-T)} & \textbf{88.7} & \textbf{40.4} & \textbf{63.2} & \textbf{45.6} \\
    \bottomrule
    \bottomrule
    \end{tabular}%
    }
      \vspace{-5pt}
  \label{t2}%
\end{table}

\begin{table}[h]
  \centering
  \vspace{-5pt}
  \caption{Per-class results (\%) on VOC to Watercolor. \textbf{The bold} represent the best results.}
    \vspace{-5pt}
   \fontsize{6}{6}\selectfont
  \resizebox{\linewidth}{!}{
    \begin{tabular}{cc|cccccc|c}
    \toprule
    \toprule
    \multicolumn{2}{c|}{\multirow{2}[4]{*}{Method}} & \multicolumn{6}{c|}{AP}                & \multicolumn{1}{c}{mAP} \\
\cmidrule{3-9}    \multicolumn{2}{c|}{} & \multicolumn{1}{c|}{Bike} & \multicolumn{1}{c|}{Bird} & \multicolumn{1}{c|}{Car} & \multicolumn{1}{c|}{Cat} & \multicolumn{1}{c|}{Dog} & \multicolumn{1}{c|}{Person} & \multicolumn{1}{c}{All} \\
    \midrule
    \multicolumn{2}{c|}{Faster R-CNN\cite{ren2015faster}} & 87.60  & 41.60  & 36.40  & 29.60  & 18.70  & 54.50  & 44.63  \\
    \multicolumn{2}{c|}{NP\cite{np}} & 87.31  & 56.22  & 50.37  & 42.05  & 41.78  & 42.18  & 53.31  \\
    \multicolumn{2}{c|}{C-Gap\cite{C_Cap}} & 86.51  & 52.59  & 49.10  & 40.55  & 39.57  & 35.68  & 50.68  \\
   \multicolumn{2}{c|}{DIV\cite{Div}} & 90.40  & 51.80  & 51.90  & 43.90  & 35.90  & \textbf{70.20}  & 57.40  \\
      \multicolumn{2}{c|}{SE-COT(Res101)\cite{SECOT}} & 91.88 & 56.74 & 50.41 & 48.22 & 41.57 & 56.19 & 57.50  \\
       \multicolumn{2}{c|}{SE-COT(Swin-T)\cite{SECOT}} & \textbf{92.69} & 63.54 & 53.97 & 48.89 & 43.13 & 61.65 & 60.70 \\
    \midrule
    \multicolumn{2}{c|}{Ours(Res101)} & 89.47 & 62.36 & \textbf{57.64} & 48.90 & 45.34 & 57.90 & 60.27 \\
    \multicolumn{2}{c|}{Ours(Swin-T)} & 90.56 & \textbf{66.70} & 52.68 & \textbf{58.20} & \textbf{47.27} & 63.78 & \textbf{63.20} \\
    
    \bottomrule
    \bottomrule
    \end{tabular}%
    }
  \label{watecolor}%
    \vspace{-5pt}
\end{table}
\textbf{Generalization from Reality to Art.} To evaluate the generalization capability of our proposed method across domains characterized by distinct stylistic shifts, we assess its effectiveness in the challenging transfer scenario from real-world images (PASCAL VOC) to artistic domains (Clipart, Watercolor, and Comic). These artistic domains exhibit significant discrepancies in texture and color distributions compared to the source domain. Table \ref{t2} presents the detection performance on these target domains. The results demonstrate that, under identical backbone configurations, our method consistently achieves superior performance across all three stylistically diverse domains. Specifically, utilizing the Swin-T backbone, our method (Swin-T) surpasses the strong baseline SE-COT by significant margins: +2.5\% mAP on Watercolor, +3.5\% mAP on Comic, and +3.1\% mAP on Clipart. Even with the ResNet-101 backbone, our method maintains a distinct advantage over SE-COT, further validating the robustness of our approach against drastic stylistic shifts.

Tables \ref{watecolor}, \ref{comic}, and \ref{clipart} provide detailed class-wise generalization results from PASCAL VOC to the Watercolor, Comic, and Clipart datasets, respectively. Notably, in critical categories such as ‘Car’ and ‘Person’, our method secures substantial performance advantages. These consistent gains confirm that our MR-DCoT framework, by simulating diverse artistic styles via the Textual Chain and enforcing structural consistency via Manifold Regression, effectively bridges the substantial domain gap between reality and art.
\begin{table*}[!t]
  \centering
   \vspace{-5pt}
  \caption{Per-class results (\%) on VOC to Clipart. \textbf{The bold sections} represent the best results.}
  \vspace{-10pt}
  \fontsize{26}{28}\selectfont
  \resizebox{\textwidth}{!}{%
    \begin{tabular}{cc|cccccccccccccccccccc|c}
    \toprule
    \toprule
    \multicolumn{2}{c|}{\multirow{2}[4]{*}{Method}} & \multicolumn{20}{c|}{AP}                                                                                                                                                                                & \multicolumn{1}{c}{mAP} \\
\cmidrule{3-23}    \multicolumn{2}{c|}{} & \multicolumn{1}{c}{Plane} & \multicolumn{1}{c}{Bike} & \multicolumn{1}{c}{Bird} & \multicolumn{1}{c}{Boat} & \multicolumn{1}{c}{Bottle} & \multicolumn{1}{c}{Bus} & \multicolumn{1}{c}{Car } & \multicolumn{1}{c}{Cat} & \multicolumn{1}{c}{Chair} & \multicolumn{1}{c}{Cow} & \multicolumn{1}{c}{Table} & \multicolumn{1}{c}{Dog} & \multicolumn{1}{c}{Horse} & \multicolumn{1}{c}{Mot.} & \multicolumn{1}{c}{Person} & \multicolumn{1}{c}{Plant} & \multicolumn{1}{c}{Sheep} & \multicolumn{1}{c}{Sofa} & \multicolumn{1}{c}{Train} & \multicolumn{1}{c|}{TV} & \multicolumn{1}{c}{All} \\
    \midrule
    \multicolumn{2}{c|}{Faster R-CNN\cite{ren2015faster}} & 34.63  & 17.02  & 28.39  & 16.37  & 13.90  & 35.51  & 44.64  & 35.14  & 19.60  & 29.05  & 20.48  & 25.70  & 39.43  & 24.41  & 34.87  & 8.98  & 17.03  & 27.78  & 29.73  & 26.96  & 26.50  \\
    \multicolumn{2}{c|}{NP\cite{np}} & 40.91  & 44.30  & 38.02  & 24.26  & 26.78  & 47.04  & 56.65  & 33.80  & 35.84  & 31.53  & 32.40  & 26.20  & 45.73  & 38.79  & 41.64  & 10.39  & 15.97  & \textbf{41.69}  & 38.17  & 37.85  & 35.40  \\
    \multicolumn{2}{c|}{C-Gap\cite{C_Cap}} & 46.08  & 42.18  & 41.04  & 24.86  & 27.25  & 43.40  & 54.60  & 39.81  & 33.27  & 40.06  & 28.50  & 32.95  & 55.32  & 38.95  & 45.03  & 10.66  & 21.54  & 37.31  & 32.68  & 38.43  & 36.74  \\
    \multicolumn{2}{c|}{DIV\cite{Div}} & 34.40 & \textbf{64.40} & 22.70 & 27.00 & \textbf{45.60} & \textbf{59.20} & 32.90 & 7.00 & \textbf{46.80} & \textbf{55.80} & 28.90 & 14.50 & 44.40 & \textbf{58.00} & \textbf{55.20} & \textbf{52.10} & 14.80 &38.40 & 42.50 & 33.90 & 38.90 \\
      \multicolumn{2}{c|}{SE-COT(Res101)\cite{SECOT}} & 51.22 & 51.86 & 50.24 & 28.93 & 26.99 & 46.23 & 55.19 & 44.46 & 33.61 & 38.26 & 34.66 & 40.22 & 59.82 & 42.82 & 49.12 & 13.13 & 25.93 & 34.75 & 37.87 & 38.29 & 40.20 \\
    \multicolumn{2}{c|}{SE-COT(Swin-T)\cite{SECOT}}&\textbf{53.48} & 54.24 & \textbf{51.87} & 31.44 & 28.35 & 49.82 & 56.02  & 46.56 & 32.06  & 40.99 & \textbf{39.24} & 43.94 & \textbf{63.10} & 46.36 & 52.42 & 14.47 & 28.95 & 36.93 & 38.04  & 41.06 & 42.50 \\

    \midrule
     \multicolumn{2}{c|}{Ours(Res101)} &48.31 & 45.87 & 39.10 & 40.20 & 26.41 & 47.89 & 55.30 & \textbf{57.12} & 46.71 & 45.61 & 31.05 & 45.38 & 52.18 & 46.32 & 50.21 & 28.15 & 28.80 & 36.59 & 43.25 & \textbf{47.75} & 43.11\\
    \multicolumn{2}{c|}{Ours(Swin-T)} & 49.32 & 50.27 & 34.87 & \textbf{43.60} & 33.79 & 50.34 & \textbf{62.89} & 56.71 & 36.20 & 46.31 & 38.76 & \textbf{51.80} & 59.90 & 47.61 & 52.48 & 31.90 & \textbf{30.45} & 41.12 & \textbf{48.30} & 46.92 & \textbf{45.58} \\
    \bottomrule
    \bottomrule
    \end{tabular}%
    }
  \label{clipart}%
   \vspace{-15pt}
\end{table*}%

\begin{table}[t]
  \centering
  \caption{Per-class results (\%) on VOC to Comic. \textbf{The bold} represent the best results.}
   \fontsize{6}{6}\selectfont
  \resizebox{\linewidth}{!}{
    \begin{tabular}{cc|cccccc|c}
    \toprule
    \toprule
    \multicolumn{2}{c|}{\multirow{2}[4]{*}{Method}} & \multicolumn{6}{c|}{AP}               & \multicolumn{1}{c}{mAP} \\
\cmidrule{3-9}    \multicolumn{2}{c|}{} & \multicolumn{1}{c|}{Bike} & \multicolumn{1}{c|}{Bird} & \multicolumn{1}{c|}{Car} & \multicolumn{1}{c|}{Cat} & \multicolumn{1}{c|}{Dog} & \multicolumn{1}{c|}{Person} & \multicolumn{1}{c}{All} \\
    \midrule
    \multicolumn{2}{c|}{Faster R-CNN\cite{ren2015faster}} & 36.50  & 8.60  & 25.90  & 9.20  & 10.80  & 25.20  & 19.37  \\
    \multicolumn{2}{c|}{NP\cite{np}} & 42.44   & 18.25  & 38.79  & 17.33  & 24.29  & 32.18  & 28.88  \\
    \multicolumn{2}{c|}{C-Gap\cite{C_Cap}} & 41.98  & 15.94  & 41.97  & 18.77  & 24.60  & 33.26 & 29.42  \\
    \multicolumn{2}{c|}{DIV\cite{Div}} & 54.10 & 16.90 & 30.10 & 25.00 & 27.40 & 45.90 & 33.20 \\
    \multicolumn{2}{c|}{YOLOv10-L \cite{yolov10}} & 51.20 & 18.40 & 31.00 & 27.40 & 28.90 & 48.40 & 34.20 \\
      \multicolumn{2}{c|}{SE-COT(Res101)\cite{SECOT}} & 47.00 & 22.99 & 44.95 & 20.91 & 29.42 & 43.63 & 34.82  \\
    \multicolumn{2}{c|}{SE-COT(Swin-T)\cite{SECOT}} & 47.73 & 21.72 & \textbf{47.60} & 23.27 & \textbf{30.78} & \textbf{50.42} & \textbf{36.92}\\
      
    \midrule
     \multicolumn{2}{c|}{Ours(Res101)} &39.56 & 27.32 & 47.87 & 24.30 & 34.62 & 50.10 & 37.30\\

    \multicolumn{2}{c|}{Ours(Swin-T)} & 42.67 & 30.40 & 48.21 & 29.30 & 36.14 & 55.80 & 40.42\\
    \bottomrule
    \bottomrule
    \end{tabular}%
    }
    \vspace{-5pt}
  \label{comic}%
\end{table}%

\begin{table}[htbp]
  \centering
  \caption{\textbf{Zero-shot domain adaptation in semantic segmentation.} Performance (mIoU\%) of our method compared against state-of-the-art approaches. Results are grouped by source domain (CS and GTA5) and target evaluation sets. CS stands for Cityscapes.}
  \label{tab:zsda_comparison_horizontal}
  \resizebox{\linewidth}{!}{
    \renewcommand{\arraystretch}{1.1} 
    \small
    \setlength{\tabcolsep}{4pt} 
    \begin{tabular}{llccccc} 
      \toprule
      \toprule
      Source & Target & Source-only & CLIPstyler \cite{Clipstyler} & P\O DA\cite{Poda} & ULDA \cite{ulda} & Ours \\
      \midrule
      \multirow{4}{*}{CS} 
        & ACDC Night & 18.31 & 21.38 & 25.03 & 25.40 & \textbf{27.20} \\
        & ACDC Snow  & 39.28 & 41.09 & 43.90 & 46.00 & \textbf{47.30} \\
        & ACDC Rain  & 38.20 & 37.17 & 42.31 & 44.94 & \textbf{45.36} \\
        & GTA5       & 39.59 & 38.73 & 40.77 & 42.91 & \textbf{44.32} \\
      \midrule
      GTA5 & Cityscapes & 36.38 & 32.40 & 40.02 & 41.73 & \textbf{43.96} \\
      \bottomrule
      \bottomrule
    \end{tabular}
  }
\end{table}
\textbf{Performance on Zero-Shot Domain Adaptation.} As presented in Table \ref{tab:zsda_comparison_horizontal}, we evaluate the generalization capability of our method on standard Zero-Shot Domain Adaptation (ZSDA) benchmarks. We benchmark our framework against the Source-only baseline and state-of-the-art approaches, including CLIPstyler \cite{Clipstyler}, P\O DA \cite{Poda}, and ULDA \cite{ulda}. In the challenging Cityscapes (CS) $\to$ ACDC scenarios, which involve continuous and severe weather shifts (Night, Snow, and Rain), our method consistently outperforms all competing approaches. Specifically, in the highly difficult Night domain where visibility is drastically reduced, our method achieves 27.20\% mIoU, surpassing the previous best method ULDA (25.40\%). Similarly, under Snow and Rain conditions, we achieve 47.30\% and 45.36\% mIoU, respectively, demonstrating the robustness of our manifold regression mechanism against severe environmental degradations. Furthermore, in the Synthetic-to-Real (GTA5 $\to$ Cityscapes) setting, which represents a substantial domain gap from virtual rendering to real-world scenes, our method attains 43.96\% mIoU. This result outperforms ULDA (41.73\%), validating that our Visual-Text Dual-CoT can effectively simulate realistic variations to bridge the domain gap without accessing any target data.

\subsection{Ablation Study}

\textbf{Component Analysis.} To validate the effectiveness of each component within our MR-DCoT framework, we conducted a comprehensive ablation study on Disentangled Feature Embedding (DFE), Visual-Text Dual Chain-of-Thought (Dual-CoT), and Manifold Regression (MR). Results are shown in Table \ref{abs}, with the first row representing the baseline model. Rows 2 and 3 apply Dual-CoT and MR on entangled features, respectively. While they offer minor improvements, their potential is constrained by the coupling of style and content, confirming the necessity of disentanglement stated in the caption. Row 5 introduces DFE combined with Dual-CoT. By explicitly decoupling style from content, Dual-CoT can simulate more effective off-manifold outliers, resulting in a significant performance leap (e.g., +9.7\% on Night Sunny compared to the baseline), which validates the visual-text chain's ability to bridge domain gaps via structured style evolution. Row 6 combines DFE with MR. While MR alone enhances stability by anchoring features to prototypes, it lacks the diverse hard samples provided by Dual-CoT, resulting in lower gains than Row 5. The final row exhibits the full model, where the synergy between Dual-CoT and MR is maximized. Dual-CoT generates challenging off-manifold samples, while MR enforces their regression back to the semantic manifold. This simulate-and-rectify mechanism achieves the best performance across all scenarios, outperforming the baseline by an average of +10.3\% on the target domains.

\begin{table}[!t]
  \centering
  \caption{\textbf{Ablation analysis of our proposed model.} DFE stands for Disentangled Feature Embedding, Dual-CoT denotes Visual-Text Dual Chain-of-Thought, and MR represents Manifold Regression.}
  \vspace{-5pt} 
  \fontsize{22}{22}\selectfont 
  \resizebox{\linewidth}{!}{
    \begin{tabular}{ccc|c|cccc}
    \toprule
    \toprule
    \multicolumn{3}{c|}{Method} & \multicolumn{1}{c|}{Source} & \multicolumn{4}{c}{Target} \\
    \midrule
    \multicolumn{1}{c}{DFE} & \multicolumn{1}{c}{Dual-CoT}  & \multicolumn{1}{c|}{MR} & \multicolumn{1}{c|}{Day Clear} &  \multicolumn{1}{c}{Night Sunny}& \multicolumn{1}{c}{Dusk Rainy} & \multicolumn{1}{c}{Night Rainy}  & \multicolumn{1}{c}{Day Foggy}  \\
    \midrule
     & &             &  58.3  & 43.4  & 40.3  & 27.5  & 36.6 \\
    
     & $\checkmark$ & & 62.7  & 47.3  & 43.5  & 30.1  & 39.2 \\

     &  & $\checkmark$ & 63.2    & 45.1  & 41.6  & 28.4  & 38.0 \\
    
    
    $\checkmark$ & $\checkmark$ & & 64.9  & 53.1  & 51.7  & 34.2  & 45.7 \\

    $\checkmark$ &  & $\checkmark$ & 63.7  & 46.2  & 43.8  & 30.2  & 42.1 \\  
    
    $\checkmark$ & $\checkmark$ & $\checkmark$ & \textbf{65.1} & \textbf{54.6} & \textbf{53.2} & \textbf{36.4} & \textbf{48.3} \\
    \bottomrule
    \bottomrule
    \end{tabular}%
  }
  \vspace{-15pt}
  \label{abs}%
\end{table}%

\textbf{Necessity Analysis of the Full Pipeline.} 
As reported in Table~\ref{tab:simplified_alternatives}, we further compare MR-DCoT with several simplified alternatives to examine whether the proposed pipeline can be replaced by simpler designs. 
The first row is the baseline detector without DFE, Dual-CoT, or MR. 
The second row introduces the stronger augmentation strategy used in DIV \cite{Div} together with consistency regularization, while the third row further applies this strategy on disentangled features. 
Although these two variants improve the baseline, they still perform clearly worse than the full MR-DCoT, indicating that stronger augmentation and consistency constraints are still insufficient to model structured and semantically controllable off-manifold shifts. 
The fourth row removes the diffusion-guided visual evolution from Dual-CoT while retaining the textual-chain guidance and MR. 
Its performance drop, especially on Dusk Rainy and Day Foggy, shows that diffusion-based local structural perturbation is important for simulating structured off-manifold degradation. 
The fifth row replaces prototype-anchored MR with prototype-free L2 regression, where off-manifold features are only regressed to their paired clean features without class-level prototype constraints. 
This variant suffers a clear degradation, particularly under challenging rainy conditions, suggesting that instance-level regression alone lacks stable class-discriminative guidance. 
Overall, the full MR-DCoT achieves the best performance across all target domains, demonstrating that the proposed pipeline is not merely a complex combination of modules, but relies on the synergy between structured off-manifold simulation and prototype-anchored manifold regression.

\begin{table}[t]
  \centering
  \caption{\textbf{Ablation analysis of simplified alternatives.} We compare MR-DCoT with stronger augmentation, no-diffusion variants, and prototype-free regression.}
  \vspace{-5pt} 
  \fontsize{22}{22}\selectfont
  \resizebox{\linewidth}{!}{
    \begin{tabular}{l|cccc} 
    \toprule
    \toprule
    \multirow{2}{*}{\textbf{Method}} 
    & \multicolumn{4}{c}{\textbf{Target Domains (mAP \%)}} \\
    \cmidrule(lr){2-5}
    & \textbf{Night Sunny} & \textbf{Dusk Rainy} & \textbf{Night Rainy} & \textbf{Day Foggy} \\
    \midrule
    Baseline & 43.4 & 40.3 & 27.5 & 36.6 \\
    StrongAug. + Cons. & 48.5 & 45.2 & 33.1 & 42.8 \\
    DFE + StrongAug. + Cons. & 49.8 & 46.7 & 35.4 & 46.5 \\
    DFE + Dual-CoT w/o Diffusion + MR & 52.7 & 49.5 & 33.7 & 44.9 \\
    DFE + Dual-CoT + L2 Reg. w/o Proto. & 50.9 & 47.2 & 30.1 & 45.4 \\
    \textbf{DFE + Dual-CoT + MR} & \textbf{54.6} & \textbf{53.2} & \textbf{36.4} & \textbf{48.3} \\
    \bottomrule
    \bottomrule
    \end{tabular}%
  }
  \vspace{-10pt}
  \label{tab:simplified_alternatives}
\end{table}

\begin{table}[h]
  \centering
  \caption{\textbf{Internal ablation of the Visual-Text Dual-CoT module.} We investigate the synergies between global semantic guidance and local structural perturbation. }
  \vspace{-5pt} 
  \fontsize{22}{22}\selectfont
  \resizebox{\linewidth}{!}{
    \begin{tabular}{cc|cccc} 
    \toprule
    \toprule
    \multicolumn{2}{c|}{Dual-CoT Components} & \multicolumn{4}{c}{Target Domains (mAP \%)} \\
    \midrule
    \multicolumn{1}{c}{Textual Chain} & \multicolumn{1}{c|}{Visual Chain} & \multicolumn{1}{c}{Night Sunny} & \multicolumn{1}{c}{Dusk Rainy} & \multicolumn{1}{c}{Night Rainy} & \multicolumn{1}{c}{Day Foggy} \\
    \midrule
    
    \checkmark &  & 52.7 & 49.5 &33.7 & 44.9 \\
    
     & \checkmark & 51.7 & 45.9 & 32.5 & 45.3 \\
    
    \checkmark & \checkmark &  \textbf{54.6} & \textbf{53.2} & \textbf{36.4} & \textbf{48.3}  \\
    \bottomrule
    \bottomrule
    \end{tabular}%
  }

  \label{tab:dual_cot_ablation}
\end{table}
\textbf{Internal Analysis of Visual-Text Dual-CoT.} As reported in Table \ref{tab:dual_cot_ablation}, we further investigate the specific contributions and synergies of the two chains within the Dual-CoT module. The first row evaluates the Textual Chain in isolation. It demonstrates robust performance in scenarios dominated by global illumination shifts, such as Night Sunny (52.7\%), validating its capability to simulate macroscopic style variations via semantic-driven AdaIN. The second row assesses the Visual Chain alone. Interestingly, it rivals or even slightly outperforms the Textual Chain in the Day Foggy domain (45.3\% vs. 44.9\%), as the diffusion-based blurring effectively mimics the structural degradation and loss of high-frequency details inherent to foggy conditions. The final row highlights the superior performance of the Full Dual-CoT. This combination proves critical for compound domains like Dusk Rainy, boosting performance to 53.2\%, a significant improvement over either chain used in isolation. This confirms that effectively handling complex real-world shifts requires both global semantic guidance and local structural perturbation to generate comprehensive off-manifold samples.

\begin{table}[t]
\centering
\caption{Model Comparison in Terms of mAP, Parameters, FPS, FLOPs, Inference Time, and GPU Memory. The mAP refers to the average mAP (\%) across the four target domains. 
Memory denotes peak training GPU memory per RTX 4090.}
\vspace{-5pt}
\fontsize{12}{13}\selectfont
\resizebox{1.0\linewidth}{!}{
\begin{tabular}{lcccccc}
\toprule
\toprule
\textbf{Model} & \textbf{mAP (\%)} & \textbf{Params (M)} & \textbf{FPS} & \textbf{FLOPs (G)} 
& \textbf{Infer. (ms)} 
& \textbf{Mem. (GB)} \\
\midrule
Faster R-CNN \cite{ren2015faster} & 26.2  & 60.6   & 12.4   & 267 
& 80.6 & 15.4 \\
S-DGOD \cite{S_DG}                & 28.7  & 72.4   & 10.1   & 284 
& 99.0 & 16.7 \\
C-Gap \cite{C_Cap}                & 31.6  & 140.2  & 5.5    & 423 
& 181.8 & 21.4 \\
PDOC \cite{pdoc}                  & 32.6  & 65.8   & 9.8    & 305 
& 102.0 & 22.5 \\
DIV \cite{Div}                    & 34.5  & 63.1   & 10.9   & 281 
& 91.7 & 22.6 \\
\midrule
Baseline                          & 26.2  & 60.6   & 12.4   & 267 
& 80.6 & 18.5 \\
+ DFE                             & 27.3  & 61.2   & 12.1   & 281 
& 82.6 & 19.3 \\
+ DFE + Dual-CoT                  & 36.6  & 61.2   & 12.1   & 281 
& 82.6 & 22.7 \\ 
\textbf{+ DFE + Dual-CoT + MR}    & \textbf{40.1} & 62.4 & 10.8 & 290
& 92.6 & 23.0 \\
\bottomrule
\bottomrule
\end{tabular}
}
\label{complexity}%
\end{table}

\textbf{Complexity and Module Cost Analysis.}
We compare the computational efficiency and model complexity of existing methods in Table~\ref{complexity}.
To ensure a fair comparison, all evaluated methods employ the same detection framework (Faster R-CNN) and backbone network (ResNet-101).
Beyond FPS and FLOPs, we further report inference latency and peak GPU memory usage to provide a more comprehensive cost analysis.
Compared to the baseline, the Disentangled Feature Embedding (DFE) module introduces only marginal computational overhead.
Notably, since the Dual-CoT module operates exclusively during training, it enhances feature robustness without introducing additional inference cost.
Its extra cost is mainly reflected in training memory, as structured off-manifold features need to be generated and maintained during optimization.
Finally, while Manifold Regression (MR) introduces a moderate overhead due to prototype-based feature enhancement, the full model still maintains competitive efficiency and achieves a favorable trade-off between generalization performance and computational cost.

\begin{table}[h]
\centering
\caption{Ablation study on prototype number and allocation strategy.}
\label{tab:proto_number}
\resizebox{\linewidth}{!}{
\begin{tabular}{l|c|cc}
\hline
\hline
\textbf{Prototype Setting} & \textbf{Allocation} & \textbf{Dusk Rainy} & \textbf{Night Rainy} \\
\hline
1 proto / class  & Uniform     & \textbf{53.2} & \textbf{36.4} \\
2 protos / class & Uniform     & 51.7 & 35.0 \\
3 protos / class & Uniform     & 50.9 & 34.7 \\
\hline
Fixed-10 protos  & Non-uniform & 52.1 & 35.8 \\
Fixed-12 protos  & Non-uniform & 51.5 & 35.2 \\
Fixed-15 protos  & Non-uniform & 51.3 & 34.7\\
\hline
\hline
\end{tabular}
}
\end{table}

\textbf{Analysis of Prototype Number and Allocation Strategy.}
We further evaluate different prototype configurations on two challenging unseen domains, i.e., Dusk Rainy and Night Rainy, as shown in Table~\ref{tab:proto_number}. 
The number of object classes is determined by the dataset protocol, while Table~\ref{tab:proto_number} studies different prototype configurations under the fixed class set.
In the uniform setting, each category is assigned the same number of prototypes, while in the non-uniform setting, a fixed total number of prototypes is allocated in a class-wise manner according to intra-class feature variance, with each category having at least one prototype. 
The results show that the default one-prototype-per-class setting achieves the best performance, indicating that a compact class-level prototype already provides a stable semantic anchor for manifold regression. 
Increasing the number of prototypes leads to performance degradation, suggesting that overly fine-grained partitioning may introduce redundant or unstable anchors, especially for categories with limited or long-tail samples. 
Although non-uniform allocation slightly alleviates this issue, it still underperforms the default setting. 
Therefore, we adopt one prototype per category as the default setting, which provides a stable and compact semantic anchor for manifold regression.

\begin{figure}[!t]
  \centering
  \includegraphics[width=1.0\columnwidth]{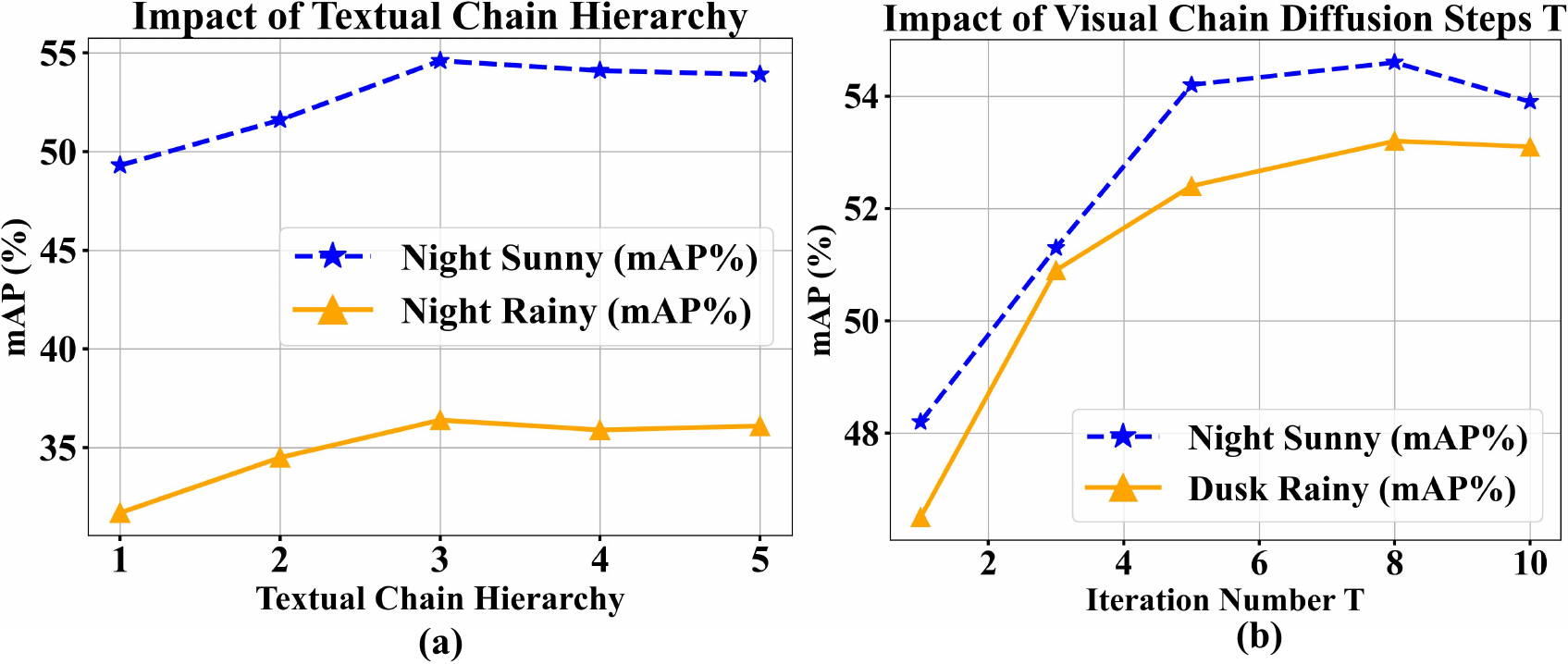}
  \vspace{-15pt}
  \caption{\textbf{Hyperparameter sensitivity analysis.} (a) \textbf{Textual Chain Hierarchy}: Performance peaks at the sentence level (Level 3), indicating an optimal balance of semantic guidance. (b) \textbf{Visual Chain Diffusion Steps $T$}: Detection accuracy increases with $T$ until saturation at $T=8$, demonstrating the trade-off between style shift magnitude and semantic preservation.}
    \label{hyper}
    \vspace{-5pt}
\end{figure}
\begin{figure}[t]
  \centering
  \includegraphics[width=1.0\columnwidth]{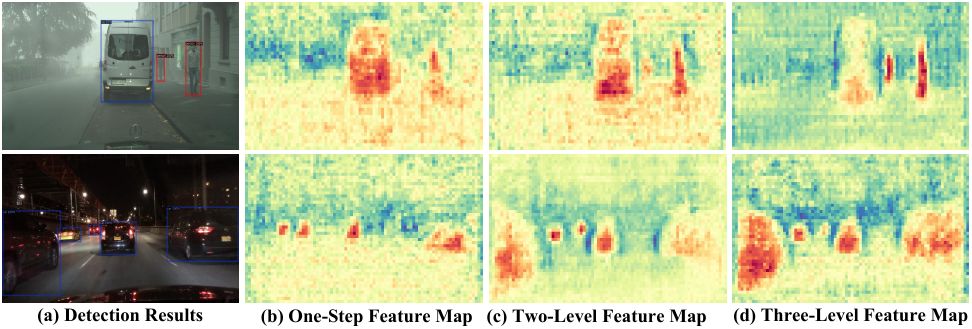}
  \vspace{-0.2in}
  \caption{\textbf{Qualitative analysis of Textual Chain Hierarchy.} We visualize the feature maps guided by different levels of the textual chain. Compared to the One-Step prompt (b), the Three-Level (Sentence-level) chain (d) significantly suppresses background noise and focuses activation on the objects, validating the effectiveness of progressive semantic guidance. }
    \label{COTF-vis}
     \vspace{-0.2in}
\end{figure}

\textbf{Impact of Textual Chain of Thought Hierarchy.} To determine the optimal granularity for style simulation, we conducted an ablation study on the hierarchical levels of the Textual Chain, ranging from discrete keywords (Level 1) to complex narrative descriptions (Level 5). As visualized in Fig. \ref{hyper}(a), the performance curves exhibit a clear ascent followed by a plateau. Initially, using simple keywords yields suboptimal results (e.g., roughly 31.7\% on Night Rainy) due to a lack of contextual richness for simulating realistic domain shifts. The performance peaks at Level 3, which corresponds to coherent sentence-level descriptions, achieving the best mAPs of 36.4\% and 54.6\% on Night Rainy and Night Sunny, respectively. This indicates that sentence-level prompts provide the most effective balance of semantic guidance. However, as the complexity further increases to Levels 4 and 5, the curves show saturation or slight degradation. We hypothesize that overly complex descriptions may introduce semantic noise or irrelevant details that distract the style evolution process. Consequently, we adopt Level 3 as the optimal setting for our Textual Chain. Simultaneously, we present a feature visualization analysis of the hierarchical Chain-of-Thought in Fig. \ref{COTF-vis}.

\textbf{Prompt Robustness and Leakage-Control Analysis.}
To examine whether the textual chain is sensitive to LLM-generated prompts or potential target-domain word leakage, we conduct a control experiment on two weather-related target domains in Table~\ref{tab:prompt_robustness}. In the default setting, the keyword groups are generated offline by GPT-4 and fixed throughout training and evaluation. Compared with this default setting, using GPT-5-regenerated keyword groups or Gemini-generated keyword groups leads to only minor performance changes, indicating that the proposed method is not tied to a specific LLM version or generated vocabulary. Moreover, when explicit weather-related words such as ‘rainy’ and ‘foggy’ are replaced with generic degradation descriptions, e.g., ‘visibility degradation with water-like streaks’ and ‘reduced contrast with low visibility’, the model still maintains competitive performance. This suggests that the improvement does not come from memorizing target-domain names, but from learning category-preserving semantic directions for domain variation.

\begin{table}[t]
\centering
\caption{\textbf{Prompt robustness and leakage-control analysis.} 
We evaluate the influence of different LLM-generated keyword groups and explicit weather-related words.}
\vspace{-5pt}
\fontsize{12}{13}\selectfont
\resizebox{\linewidth}{!}{
\begin{tabular}{l|cc}
\toprule
\toprule
\textbf{Prompt Setting} & \textbf{Dusk Rainy} & \textbf{Day Foggy} \\
\midrule
Fixed offline keyword groups & 53.2 & 48.3 \\
GPT-5 re-generated keyword groups \cite{gpt-5} & \textbf{53.3} & 47.9 \\
Gemini-generated keyword groups\cite{gemini} & 53.0 & 48.1 \\
Generic descriptions w/o explicit weather words & 52.8 & \textbf{48.5} \\
\bottomrule
\bottomrule
\end{tabular}
}
\vspace{-10pt}
\label{tab:prompt_robustness}
\end{table}

\begin{table}[!h]
  \centering
  \vspace{-10pt}
  \caption{\textbf{Sensitivity analysis of sampled keyword number.} 
  We evaluate different numbers of keywords used to construct the Level-1 textual feature.}
  \label{tab:keyword_number_sensitivity}
  
  \renewcommand{\arraystretch}{1.2}
  \setlength{\tabcolsep}{5pt}       
  \fontsize{12}{5}\selectfont     
  \resizebox{1.0\linewidth}{!}{
    \begin{tabular}{cccc|cc}
    \toprule
    \toprule
    \multicolumn{4}{c|}{\textbf{Keyword Number}} & \multicolumn{2}{c}{\textbf{Target Domains (mAP \%)}} \\
    \cmidrule(r){1-4} \cmidrule(l){5-6}
    \textbf{$m_r=3$} & \textbf{$m_r=5$} & \textbf{$m_r=6$} & \textbf{$m_r=7$} & \textbf{Dusk Rainy} & \textbf{Day Foggy} \\
    \midrule

    \checkmark & & & & 52.9 & 47.8 \\

    & \checkmark & & & \textbf{53.2} & 48.3 \\

    & & \checkmark & & 53.0 & \textbf{48.5} \\

    & & & \checkmark & 53.1 & 47.9 \\

    \bottomrule
    \bottomrule
    \end{tabular}%
  }
  \vspace{-5pt}
\end{table}

\textbf{Sensitivity to Sampled Keyword Number.}
As shown in Table~\ref{tab:keyword_number_sensitivity}, we further analyze the influence of the sampled keyword number $m_r$ in the Textual Chain. 
The results show that the performance remains stable when $m_r$ varies from 3 to 7, indicating that our method is not sensitive to this hyperparameter. 
When $m_r$ is relatively small, the textual guidance may provide insufficient semantic diversity, while using more keywords may introduce redundant or weakly relevant descriptions. 
Overall, $m_r=5$ achieves the best result on Dusk Rainy and competitive performance on Day Foggy, providing a favorable balance between semantic diversity and textual controllability. 
Therefore, we set $m_r=5$ as the default setting in our experiments.

\begin{table}[!t]
  \centering
  \vspace{-5pt}
  \caption{\textbf{Impact of Injection Layers for Off-Manifold Generation.} We investigate applying the Visual-Text Dual-CoT at different feature levels. }
  \label{layer}
  \vspace{-5pt}
  \renewcommand{\arraystretch}{1.2}
  \setlength{\tabcolsep}{2pt}       
  \fontsize{12}{5}\selectfont     
  \resizebox{1.0\linewidth}{!}{
    \begin{tabular}{cccc|cc}
    \toprule
    \toprule
    \multicolumn{4}{c|}{Injection Layer} & \multicolumn{2}{c}{Target Domains (mAP \%)} \\
    \cmidrule(r){1-4} \cmidrule(l){5-6}
    Layer 1 & Layer 2 & Layer 3 & Layer 4 & Night Rainy & Night Sunny \\
    \midrule

    \checkmark & & & & \textbf{36.4} & \textbf{54.6} \\

     & \checkmark & & & 34.3 & 52.1 \\ 

     & & \checkmark & & 30.6 & 47.2 \\ 
    
     & & & \checkmark & 30.2 & 46.4 \\ 
     
    \bottomrule
    \bottomrule
    \end{tabular}%
    
  }
  \vspace{-15pt}
\end{table}
\textbf{Impact of Visual Chain Diffusion Steps ($T$).}
We further investigate the impact of the diffusion iteration number $T$ within the Visual Chain, which explicitly controls the magnitude of local structural perturbations.
As illustrated in Fig. \ref{hyper}(b), increasing $T$ from 1 to 8 significantly boosts detection accuracy, with Night Sunny and Dusk Rainy peaking at 54.6\% and 53.2\%, respectively. This confirms that a sufficient diffusion depth is necessary to generate hard off-manifold samples, forcing the model to learn structural invariance against severe visual corruptions.
However, pushing the diffusion steps further to $T=10$ results in a slight performance degradation. We attribute this to semantic destruction: excessive diffusion irreversibly damages critical object details, making it difficult for the regression module to anchor features back to the source prototypes. Consequently, we identify $T=8$ as the optimal setting to balance perturbation strength and semantic preservation. 
Simultaneously, we present the t-SNE visualization of off-manifold simulation and manifold regression in Fig. \ref{fig:tsne_evolution}.

\begin{figure*}[t]
  \centering
  \includegraphics[width=1.0\linewidth]{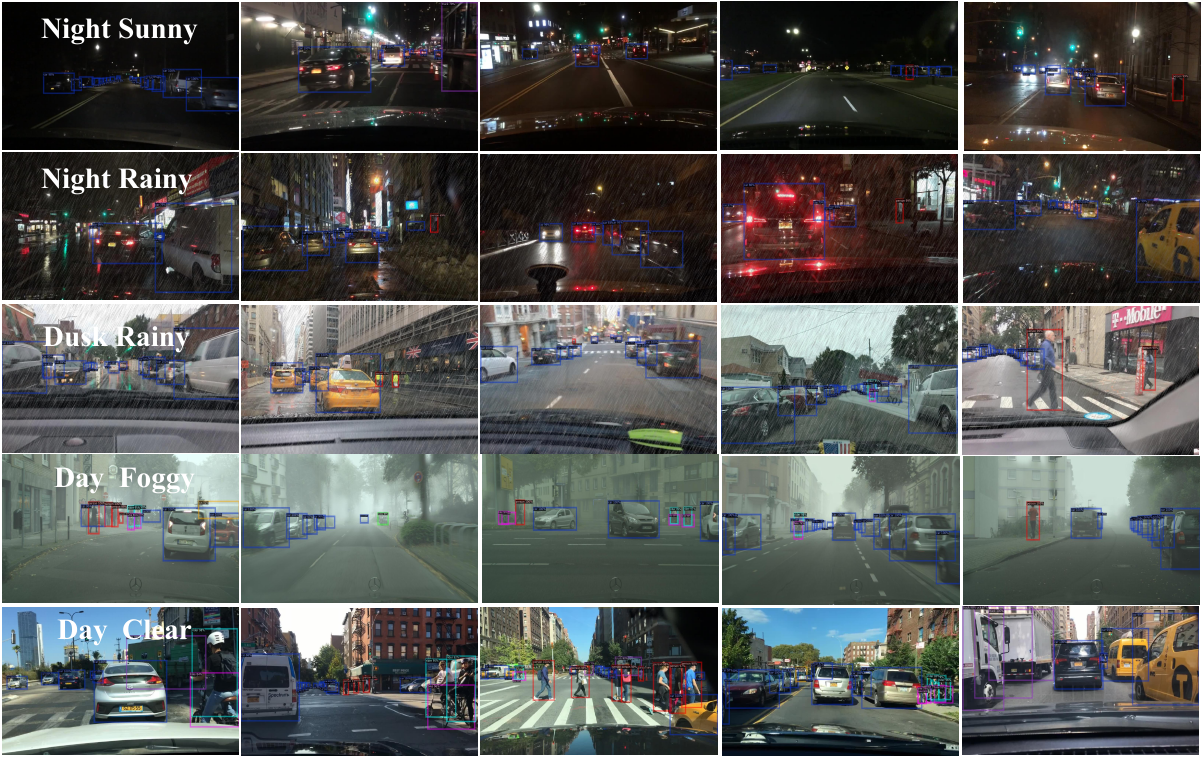} 
   \vspace{-20pt}
  \caption{\textbf{Qualitative detection results across five distinct weather scenarios.} We showcase the inference performance on the source domain (Day Clear) and four unseen target domains (Night Sunny, Night Rainy, Dusk Rainy, and Day Foggy). Despite severe illumination changes, rain streak occlusions, and low-contrast conditions, our method produces precise bounding boxes with high confidence, demonstrating robust generalization capabilities.}
  \label{fig:vis_all}
  \vspace{-10pt}
\end{figure*}

\textbf{Impact of Injection Layers.} To determine the optimal depth for feature evolution, we applied the Visual-Text Dual-CoT module across different feature levels of the backbone (Layer 1 to Layer 4). As shown in Table \ref{layer}, the best performance is achieved when injecting perturbations at Layer 1, yielding mAPs of 36.4\% and 54.6\% on Night Rainy and Night Sunny, respectively. Performance degrades significantly as the injection position shifts to deeper layers (e.g., Layer 4 drops to 30.2\% on Night Rainy). We attribute this to the nature of feature hierarchies: shallow layers (Layer 1) primarily encode low-level statistics such as color, texture, and lighting, which correspond to style variations. Modulating these layers effectively simulates domain shifts while preserving the underlying object structure. In contrast, deep layers encode high-level semantic abstractions. Directly perturbing these layers risks corrupting the semantic identity of objects, thereby hindering the Manifold Regression module from recovering the correct class prototypes. Consequently, we select Layer 1 as the optimal injection level.

\begin{figure}[!t]
  \centering
  \includegraphics[width=1.0\columnwidth]{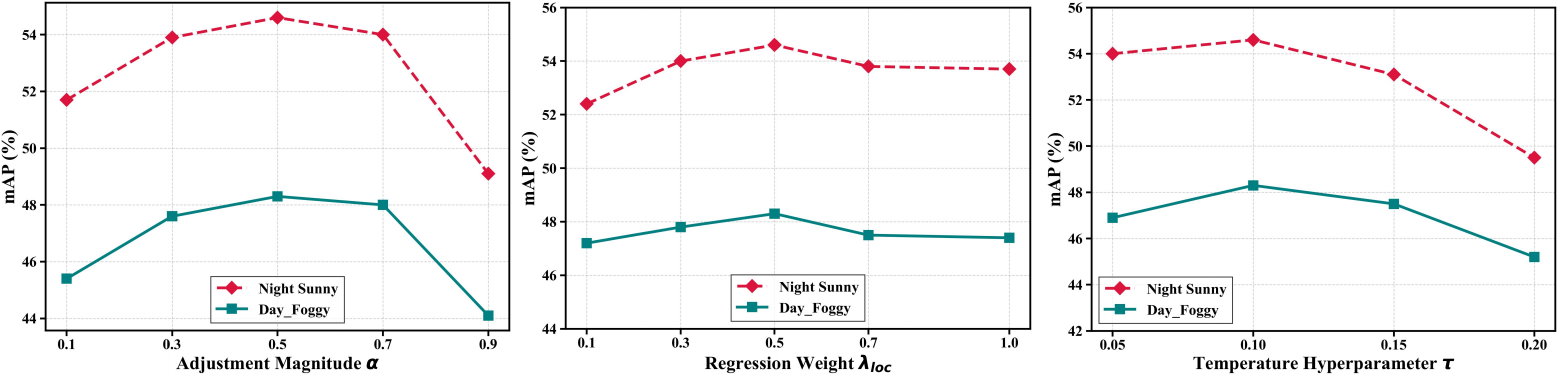}
  \vspace{-15pt}
  \caption{\textbf{More Hyperparameter Sensitivity Analysis.} The impact of dynamic blur magnitude $\alpha$, regression weight $\lambda_{loc}$, and temperature hyperparameter $\tau$ on detection performance (mAP) across Night Sunny and Day Foggy domains.}
    \label{hyper2}
    \vspace{-15pt}
\end{figure}
\noindent\textbf{More Hyperparameter Sensitivity Analysis.} 
We further investigate the sensitivity of MR-DCoT to three key hyperparameters, as illustrated in Fig.~\ref{hyper2}. 
First, for the dynamic blur magnitude $\alpha$, the detection performance exhibits an inverted U-shaped trend, peaking at $\alpha=0.5$. Lower values fail to generate sufficiently challenging outliers, while excessive blurring ($\alpha=0.9$) destroys essential structural cues. 
Second, the regression weight $\lambda_{loc}$ achieves optimality at 0.5, confirming that a balanced trade-off between global semantic anchoring and local structural rectification is crucial for robust rectification. 
Finally, the temperature $\tau$ in the contrastive objective is highly sensitive; setting $\tau=0.1$ ensures appropriate manifold concentration, whereas larger values overly smooth the distribution, significantly reducing class discriminability.

\begin{figure*}[t]
  \centering
  \includegraphics[width=2.0\columnwidth]{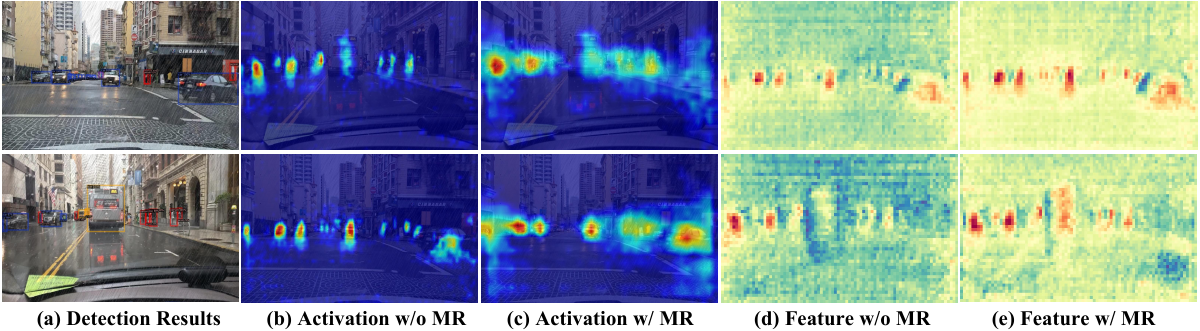}
  \vspace{-0.1in}
  \caption{\textbf{Qualitative analysis of Manifold Regression.} We visualize the activation and feature maps to evaluate the rectification capability. Compared to the baseline without Manifold Regression (b, d), our method (c, e) significantly suppresses background noise (e.g., rain streaks and road reflections) and focuses activation on the semantic objects (vehicles), validating the effectiveness of our prototype-anchored rectification mechanism. }
    \label{f-vis}
     \vspace{-0.15in}
\end{figure*}

\subsection{Visualization Analysis}
\textbf{Qualitative Analysis on Diverse Scenarios.} To intuitively evaluate the generalization boundaries of our method, we present comprehensive detection results across five distinct weather scenarios in Fig. \ref{fig:vis_all}. These visualizations demonstrate the model's robustness against various environmental corruptions.
First, regarding illumination robustness (Night Sunny/Night Rainy), as illustrated in the first and second rows, night scenes are plagued by extremely low visibility and interference from street lamps and vehicle glare. Despite these challenges, our model successfully suppresses glare and accurately detects objects within dark regions, validating the generalization capability of our method under conditions of extremely low visibility and intense glare.
Second, concerning weather artifact resilience (Rainy/Dusk), in the Dusk Rainy and Night Rainy scenarios (second and third rows), images are severely degraded by dense rain streaks and water reflections on the windshield. Our method exhibits remarkable resilience to such high-frequency noise, effectively avoiding false positives caused by reflections while maintaining high recall for occluded vehicles.
Finally, regarding small object detection, it is worth noting that even in complex scenarios with dense traffic (e.g., the crowded street shown in the bottom rows), our model precisely localizes small-scale and distant objects. In summary, these qualitative results substantiate the robustness of our method across diverse generalization scenarios and against drastic domain shifts.

\textbf{Visualization of Manifold Regression Impact.} To intuitively validate the effectiveness of our Manifold Regression (MR) module, we present a comparative visualization of feature responses with and without MR in Fig. \ref{f-vis}. As illustrated in the baseline configuration without MR (Fig. \ref{f-vis}b, d), the activation patterns are scattered and heavily distracted by environmental noise, such as rain streaks and wet road reflections. This indicates that the features have significantly deviated from the underlying semantic manifold. In contrast, incorporating Manifold Regression (Fig. \ref{f-vis}c, e) yields significantly more compact and precise activation maps. The model successfully suppresses style-induced noise and refocuses attention intensely on the semantic objects (e.g., vehicles). This visual evidence aligns perfectly with our proposed Regress-to-Rectify paradigm, confirming that anchoring deviant features back to class-specific prototypes is essential for preserving semantic consistency under severe structural degradations.

\textbf{Visualization of Textual Chain Hierarchy Impact.} To intuitively validate the influence of the Textual Chain Hierarchy on feature learning, we present feature map visualizations in Fig. \ref{COTF-vis} under foggy and night scenarios. These comparisons reveal a clear progression in feature discriminability. In the One-Step setting (Fig. \ref{COTF-vis}b), representative of methods like C-Gap, feature activations appear scattered and heavily contaminated by environmental interference, such as thick fog in foggy scenes or uneven illumination in night scenes. This confirms that a single-step prompting strategy is insufficient for guiding the model to distinguish semantic information from complex environmental style shifts. As the hierarchy deepens (Fig. \ref{COTF-vis}c), background interference is gradually suppressed. Finally, the Three-Level setting (Fig. \ref{COTF-vis}d), corresponding to the optimal sentence-level guidance, exhibits the most compact and precise activation maps. The model successfully suppresses style-induced noise and refocuses intensely on semantic objects (e.g., vehicles and pedestrians). This visual evidence aligns perfectly with our quantitative results, demonstrating that coherent sentence-level descriptions provide the robust semantic anchors necessary for resisting severe domain shifts.

\begin{figure}[t]
  \centering
  \includegraphics[width=1.0\linewidth]{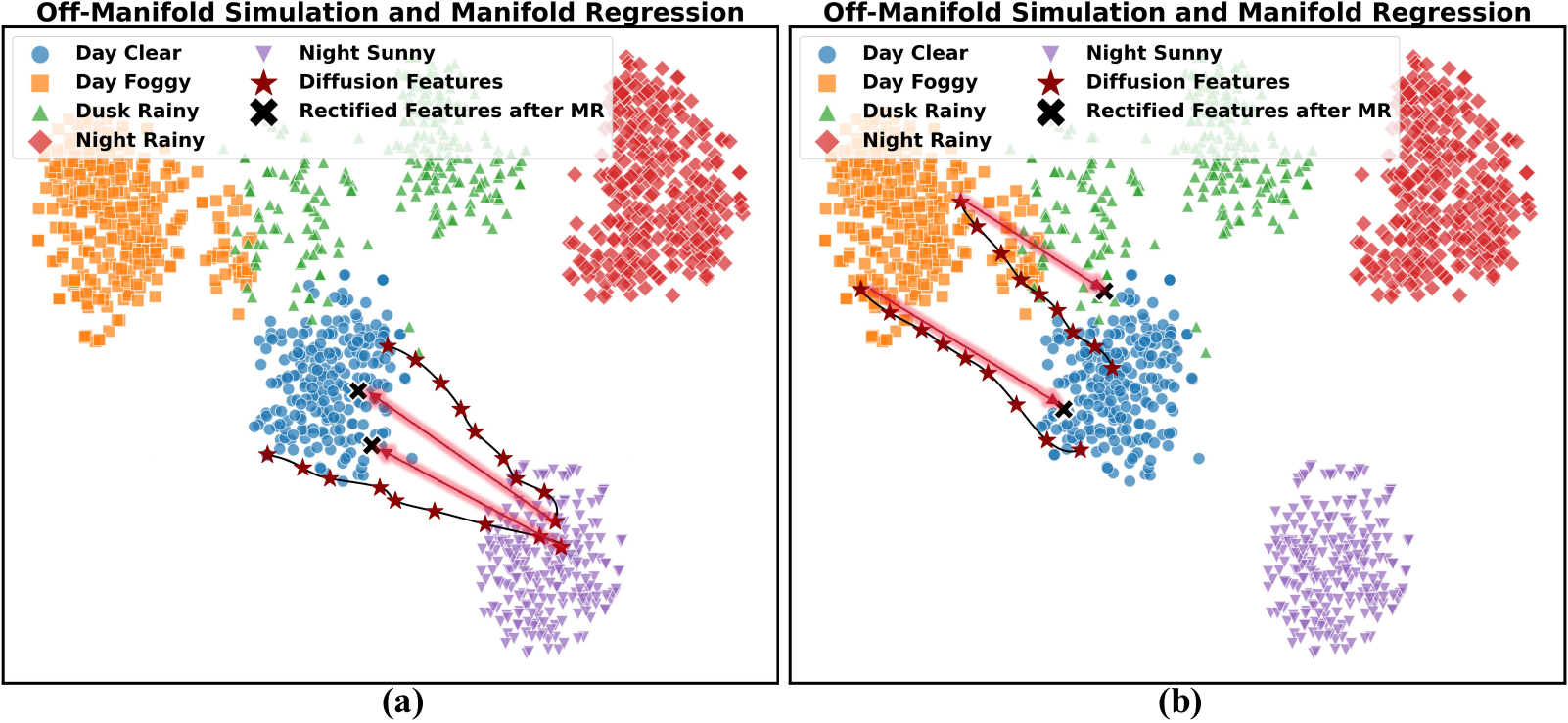}
  \vspace{-15pt}
  \caption{\textbf{t-SNE Visualization of Off-Manifold Simulation and Manifold Regression.} 
  The two subplots show that Dual-CoT generates off-manifold features under different textual guidance, while Manifold Regression rectifies these deviant features back toward the source semantic region.}
  \vspace{-5pt}
  \label{fig:tsne_evolution}
\end{figure}

\textbf{Visualization Analysis of Off-Manifold Simulation and Manifold Regression.}
To validate the effectiveness of the proposed Dual-CoT and Manifold Regression modules, we visualize the feature evolution process in the t-SNE space.
As illustrated in Fig. \ref{fig:tsne_evolution}, both settings start from the source-domain distribution, i.e., Day Clear, and use different textual descriptions to guide the direction of off-manifold simulation.
When the textual prompt is set to ‘Driving at night’, the Dual-CoT features progressively drift toward the Night Sunny distribution, while the prompt ‘Driving on a Foggy day’ drives the generated features toward the Day Foggy distribution.
This demonstrates that Dual-CoT can produce controllable off-manifold feature trajectories under semantic guidance, thereby effectively expanding the source-domain feature distribution.
Furthermore, we visualize the features after Manifold Regression, marked as black crosses in Fig. \ref{fig:tsne_evolution}.
Compared with the Dual-CoT-induced off-manifold features, the rectified features move back toward the source-domain semantic region, indicating that Manifold Regression can correct shift-induced feature deviations rather than merely generating stronger perturbations.
This visualization provides intuitive evidence for the proposed simulate-to-rectify paradigm: Dual-CoT first pushes source features away from the source semantic support to construct challenging off-manifold samples, while Manifold Regression subsequently guides these deviant features back to the source semantic neighborhood.

\section{Conclusion}
For the Single-DGOD task, we propose a new method, i.e., Manifold Regression with Visual-Text Dual Chain-of-Thought (MR-DCoT), which rethinks robust generalization as a manifold regression problem rather than simple data simulation. This approach utilizes a Visual-Text Dual Chain-of-Thought to construct structured and high-difficulty off-manifold samples, where the textual chain specifies controllable global semantic directions and the visual chain injects fine-grained structural perturbations. Additionally, we introduce a Class-Specific Prototype Anchoring mechanism to guide deviant features back toward the source semantic manifold, thereby equipping the model with a stable error-correction capability. Experimental results demonstrate that our method outperforms state-of-the-art approaches across diverse adverse weather scenarios, Real-to-Art benchmarks, and zero-shot semantic segmentation tasks, validating the versatility and effectiveness of the proposed Simulation-Regression paradigm.

\bibliographystyle{IEEEtran}
\bibliography{sample-base.bib}

@String(CVPR= {IEEE Conf. Comput. Vis. Pattern Recog.})

@String(AAAI = {AAAI})

@String(CVPR  = {CVPR})

@article{VOC,
  title={The pascal visual object classes (voc) challenge},
  author={Everingham, Mark and Van Gool, Luc and Williams, Christopher KI and Winn, John and Zisserman, Andrew},
  journal={International Journal of Computer Vision},
  volume={88},
  pages={303--338},
  year={2010},
  publisher={Springer}
}

@article{sfda,
  title={A comprehensive survey on source-free domain adaptation},
  author={Li, Jingjing and Yu, Zhiqi and Du, Zhekai and Zhu, Lei and Shen, Heng Tao},
  journal={IEEE Transactions on Pattern Analysis and Machine Intelligence},
  year={2024},
  publisher={IEEE}
}

@article{pamidoj,
  title={Robust domain adaptive object detection with unified multi-granularity alignment},
  author={Zhang, Libo and Zhou, Wenzhang and Fan, Heng and Luo, Tiejian and Ling, Haibin},
  journal={IEEE Transactions on Pattern Analysis and Machine Intelligence},
  year={2024},
  publisher={IEEE}
}

@ARTICLE{Diffs,
  author={Zhang, Mingyuan and Cai, Zhongang and Pan, Liang and Hong, Fangzhou and Guo, Xinying and Yang, Lei and Liu, Ziwei},
  journal={IEEE Transactions on Pattern Analysis and Machine Intelligence}, 
  title={MotionDiffuse: Text-Driven Human Motion Generation With Diffusion Model}, 
  year={2024},
  volume={46},
  number={6},
  pages={4115-4128},
  doi={10.1109/TPAMI.2024.3355414}}

@article{TIB,
  title={TIB: Detecting unknown objects via two-stream information bottleneck},
  author={Wu, Aming and Deng, Cheng},
  journal={IEEE Transactions on Pattern Analysis and Machine Intelligence},
  volume={46},
  number={1},
  pages={611--625},
  year={2023},
  publisher={IEEE}
}

@inproceedings{Art,
  title={Cross-domain weakly-supervised object detection through progressive domain adaptation},
  author={Inoue, Naoto and Furuta, Ryosuke and Yamasaki, Toshihiko and Aizawa, Kiyoharu},
  booktitle={Proceedings of the IEEE Conference on Computer Vision and Pattern Recognition},
  pages={5001--5009},
  year={2018}
}

@inproceedings{UFR,
  title={Unbiased Faster R-CNN for Single-source Domain Generalized Object Detection},
  author={Liu, Yajing and Zhou, Shijun and Liu, Xiyao and Hao, Chunhui and Fan, Baojie and Tian, Jiandong},
  booktitle={Proceedings of the IEEE/CVF Conference on Computer Vision and Pattern Recognition},
  pages={28838--28847},
  year={2024}
}

@inproceedings{np,
  title={Towards robust object detection invariant to real-world domain shifts},
  author={Fan, Qi and Segu, Mattia and Tai, Yu-Wing and Yu, Fisher and Tang, Chi-Keung and Schiele, Bernt and Dai, Dengxin},
  booktitle={The Eleventh International Conference on Learning Representations},
  year={2023},
  organization={OpenReview}
}

@inproceedings{Div,
  title={Improving Single Domain-Generalized Object Detection: A Focus on Diversification and Alignment},
  author={Danish, Muhammad Sohail and Khan, Muhammad Haris and Munir, Muhammad Akhtar and Sarfraz, M Saquib and Ali, Mohsen},
  booktitle={Proceedings of the IEEE/CVF Conference on Computer Vision and Pattern Recognition},
  pages={17732--17742},
  year={2024}
}

@inproceedings{S_DG,
  title={Single-domain generalized object detection in urban scene via cyclic-disentangled self-distillation},
  author={Wu, Aming and Deng, Cheng},
  booktitle={Proceedings of the IEEE/CVF Conference on Computer Vision and Pattern Recognition},
  pages={847--856},
  year={2022}
}

@inproceedings{C_Cap,
  title={Clip the gap: A single domain generalization approach for object detection},
  author={Vidit, Vidit and Engilberge, Martin and Salzmann, Mathieu},
  booktitle={Proceedings of the IEEE/CVF Conference on Computer Vision and Pattern Recognition},
  pages={3219--3229},
  year={2023}
}

@inproceedings{ISW,
  title={Robustnet: Improving domain generalization in urban-scene segmentation via instance selective whitening},
  author={Choi, Sungha and Jung, Sanghun and Yun, Huiwon and Kim, Joanne T and Kim, Seungryong and Choo, Jaegul},
  booktitle={Proceedings of the IEEE/CVF Conference on Computer Vision and Pattern Recognition},
  pages={11580--11590},
  year={2021}
}

@inproceedings{CLIP,
  title={Learning transferable visual models from natural language supervision},
  author={Radford, Alec and Kim, Jong Wook and Hallacy, Chris and Ramesh, Aditya and Goh, Gabriel and Agarwal, Sandhini and Sastry, Girish and Askell, Amanda and Mishkin, Pamela and Clark, Jack and others},
  booktitle={International Conference on Machine Learning},
  pages={8748--8763},
  year={2021},
  organization={PMLR}
}

@inproceedings{SW,
  title={Switchable whitening for deep representation learning},
  author={Pan, Xingang and Zhan, Xiaohang and Shi, Jianping and Tang, Xiaoou and Luo, Ping},
  booktitle={Proceedings of the IEEE/CVF International Conference on Computer Vision},
  pages={1863--1871},
  year={2019}
}

@article{ren2015faster,
  title={Faster r-cnn: Towards real-time object detection with region proposal networks},
  author={Ren, Shaoqing and He, Kaiming and Girshick, Ross and Sun, Jian},
  journal={Advances in Neural Information Processing Systems},
  volume={28},
  year={2015}
}

@inproceedings{Poda,
  title={PODA: Prompt-driven Zero-shot Domain Adaptation},
  author={Fahes, Mohammad and Vu, Tuan-Hung and Bursuc, Andrei and P{\'e}rez, Patrick and de Charette, Raoul},
  booktitle={Proceedings of the IEEE/CVF International Conference on Computer Vision},
  pages={18623--18633},
  year={2023}
}

@inproceedings{wu1-universal,
  title={Universal-prototype enhancing for few-shot object detection},
  author={Wu, Aming and Han, Yahong and Zhu, Linchao and Yang, Yi},
  booktitle={Proceedings of the IEEE/CVF International Conference on Computer Vision},
  pages={9567--9576},
  year={2021}
}

@inproceedings{wu2-domain,
  title={Vector-decomposed disentanglement for domain-invariant object detection},
  author={Wu, Aming and Liu, Rui and Han, Yahong and Zhu, Linchao and Yang, Yi},
  booktitle={Proceedings of the IEEE/CVF International Conference on Computer Vision},
  pages={9342--9351},
  year={2021}
}

@inproceedings{wu2025-OOD2,
  title={Percept, Memory, and Imagine: World Feature Simulating for Open-Domain Unknown Object Detection},
  author={Wu, Aming and Deng, Cheng},
  booktitle={Proceedings of the IEEE/CVF Conference on Computer Vision and Pattern Recognition (CVPR)},
  pages={4682--4691},
  year={2025}
}

@article{DFDD,
  title={Towards OOD Object Detection With Unknown-Concept Guided Feature Diffusion},
  author={Wu, Aming and Deng, Cheng},
  journal={IEEE Transactions on Pattern Analysis and Machine Intelligence},
  volume={47},
  number={11},
  pages={9798--9812},
  year={2025},
  publisher={IEEE}
}

@article{tpamidif, 
  title={Diffusion models in vision: A survey},
  author={Croitoru, Florinel-Alin and Hondru, Vlad and Ionescu, Radu Tudor and Shah, Mubarak},
  journal={IEEE Transactions on Pattern Analysis and Machine Intelligence},
  volume={45},
  number={9},
  pages={10850--10869},
  year={2023},
  publisher={IEEE}
}

@article{ScoreMatching,
  title={A connection between score matching and denoising autoencoders},
  author={Vincent, Pascal},
  journal={Neural computation},
  volume={23},
  number={7},
  pages={1661--1674},
  year={2011},
  publisher={MIT Press}
}

@inproceedings{Clipstyler,
  title={Clipstyler: Image style transfer with a single text condition},
  author={Kwon, Gihyun and Ye, Jong Chul},
  booktitle={Proceedings of the IEEE/CVF Conference on Computer Vision and Pattern Recognition},
  pages={18062--18071},
  year={2022}
}

@inproceedings{ulda,
  title={Unified language-driven zero-shot domain adaptation},
  author={Yang, Senqiao and Tian, Zhuotao and Jiang, Li and Jia, Jiaya},
  booktitle={Proceedings of the IEEE/CVF Conference on Computer Vision and Pattern Recognition},
  pages={23407--23415},
  year={2024}
}

@inproceedings{CAE,
  title={Contractive auto-encoders: Explicit invariance during feature extraction},
  author={Rifai, Salah and Vincent, Pascal and Muller, Xavier and Glorot, Xavier and Bengio, Yoshua},
  booktitle={Proceedings of the 28th international conference on international conference on machine learning},
  pages={833--840},
  year={2011}
}

@article{back,
  title={Back to Basics: Let Denoising Generative Models Denoise},
  author={Li, Tianhong and He, Kaiming},
  journal={arXiv preprint arXiv:2511.13720},
  year={2025}
}

@article{gpt,
  title={Gpt-4 technical report},
  author={Achiam, Josh and Adler, Steven and Agarwal, Sandhini and Ahmad, Lama and Akkaya, Ilge and Aleman, Florencia Leoni and Almeida, Diogo and Altenschmidt, Janko and Altman, Sam and Anadkat, Shyamal and others},
  journal={arXiv preprint arXiv:2303.08774},
  year={2023}
}

@article{gpt-5,
  title={Openai gpt-5 system card},
  author={Singh, Aaditya and Fry, Adam and Perelman, Adam and Tart, Adam and Ganesh, Adi and El-Kishky, Ahmed and McLaughlin, Aidan and Low, Aiden and Ostrow, AJ and Ananthram, Akhila and others},
  journal={arXiv preprint arXiv:2601.03267},
  year={2025}
}

@article{gemini,
  title={Gemini: a family of highly capable multimodal models},
  author={Team, Gemini and Anil, Rohan and Borgeaud, Sebastian and Alayrac, Jean-Baptiste and Yu, Jiahui and Soricut, Radu and Schalkwyk, Johan and Dai, Andrew M and Hauth, Anja and Millican, Katie and others},
  journal={arXiv preprint arXiv:2312.11805},
  year={2023}
}

@article{ada,
  title={AdaIN-based tunable CycleGAN for efficient unsupervised low-dose CT denoising},
  author={Gu, Jawook and Ye, Jong Chul},
  journal={IEEE Transactions on Computational Imaging},
  volume={7},
  pages={73--85},
  year={2021},
  publisher={IEEE}
}

@inproceedings{diffusiondet,
  title={Diffusiondet: Diffusion model for object detection},
  author={Chen, Shoufa and Sun, Peize and Song, Yibing and Luo, Ping},
  booktitle={Proceedings of the IEEE/CVF International Conference on Computer Vision},
  pages={19830--19843},
  year={2023}
}

@inproceedings{fd,
  title={Deep feature deblurring diffusion for detecting out-of-distribution objects},
  author={Wu, Aming and Chen, Da and Deng, Cheng},
  booktitle={Proceedings of the IEEE/CVF International Conference on Computer Vision},
  pages={13381--13391},
  year={2023}
}

@inproceedings{pdoc,
  title={Prompt-driven dynamic object-centric learning for single domain generalization},
  author={Li, Deng and Wu, Aming and Wang, Yaowei and Han, Yahong},
  booktitle={Proceedings of the IEEE/CVF Conference on Computer Vision and Pattern Recognition},
  pages={17606--17615},
  year={2024}
}

@inproceedings{G-NAS,
  title={G-NAS: Generalizable Neural Architecture Search for Single Domain Generalization Object Detection},
  author={Wu, Fan and Gao, Jinling and Hong, Lanqing and Wang, Xinbing and Zhou, Chenghu and Ye, Nanyang},
  booktitle={Proceedings of the AAAI Conference on Artificial Intelligence},
  volume={38},
  number={6},
  pages={5958--5966},
  year={2024}
}

@article{yolov10,
  title={Yolov10: Real-time end-to-end object detection},
  author={Wang, Ao and Chen, Hui and Liu, Lihao and Chen, Kai and Lin, Zijia and Han, Jungong and others},
  journal={Advances in Neural Information Processing Systems},
  volume={37},
  pages={107984--108011},
  year={2024}
}

@inproceedings{DGS,
  title={Deep defocus map estimation using domain adaptation},
  author={Lee, Junyong and Lee, Sungkil and Cho, Sunghyun and Lee, Seungyong},
  booktitle={Proceedings of the IEEE/CVF Conference on Computer Vision and Pattern Recognition},
  pages={12222--12230},
  year={2019}
}

@inproceedings{domainlianxu,
  title={Source-free domain adaptation with frozen multimodal foundation model},
  author={Tang, Song and Su, Wenxin and Ye, Mao and Zhu, Xiatian},
  booktitle={Proceedings of the IEEE/CVF Conference on Computer Vision and Pattern Recognition},
  pages={23711--23720},
  year={2024}
}

@article{domainshift,
  title={Behind every domain there is a shift: Adapting distortion-aware vision transformers for panoramic semantic segmentation},
  author={Zhang, Jiaming and Yang, Kailun and Shi, Hao and Rei{\ss}, Simon and Peng, Kunyu and Ma, Chaoxiang and Fu, Haodong and Torr, Philip HS and Wang, Kaiwei and Stiefelhagen, Rainer},
  journal={IEEE Transactions on Pattern Analysis and Machine Intelligence},
  year={2024},
  publisher={IEEE}
}

@inproceedings{SECOT,
  title={Style Evolving along Chain-of-Thought for Unknown-Domain Object Detection},
  author={Zhang, Zihao and Wu, Aming and Han, Yahong},
  booktitle={Proceedings of the IEEE/CVF Conference on Computer Vision and Pattern Recognition (CVPR)},
  pages={14225--14234},
  year={2025}
}

@ARTICLE{FWCL,
  author={Guo, Xiaolong and Liu, Chengxu and Qian, Xueming and Wang, Zhixiao and Feng, Xubin and Xue, Yao},
  journal={IEEE Transactions on Multimedia}, 
  title={Single-Domain Generalized Object Detection with Frequency Whitening and Contrastive Learning}, 
  year={2025},
  volume={},
  number={},
  pages={1-14},
  doi={10.1109/TMM.2025.3590915}}

@inproceedings{imagecaption,
  title={Show, attend and tell: Neural image caption generation with visual attention},
  author={Xu, Kelvin and Ba, Jimmy and Kiros, Ryan and Cho, Kyunghyun and Courville, Aaron and Salakhudinov, Ruslan and Zemel, Rich and Bengio, Yoshua},
  booktitle={International Conference on Machine Learning},
  pages={2048--2057},
  year={2015},
  organization={PMLR}
}

@inproceedings{GLIP,
  title={Grounded language-image pre-training},
  author={Li, Liunian Harold and Zhang, Pengchuan and Zhang, Haotian and Yang, Jianwei and Li, Chunyuan and Zhong, Yiwu and Wang, Lijuan and Yuan, Lu and Zhang, Lei and Hwang, Jenq-Neng and others},
  booktitle={Proceedings of the IEEE/CVF Conference on Computer Vision and Pattern Recognition},
  pages={10965--10975},
  year={2022}
}

@article{Georeg,
  title={Geodesic regression and the theory of least squares on Riemannian manifolds},
  author={Thomas Fletcher, P},
  journal={International Journal of Computer Vision},
  volume={105},
  number={2},
  pages={171--185},
  year={2013},
  publisher={Springer}
}

@inproceedings{ACDC,
  title={ACDC: The adverse conditions dataset with correspondences for semantic driving scene understanding},
  author={Sakaridis, Christos and Dai, Dengxin and Van Gool, Luc},
  booktitle={Proceedings of the IEEE/CVF International Conference on Computer Vision},
  pages={10765--10775},
  year={2021}
}

@inproceedings{GTA5,
  title={Playing for data: Ground truth from computer games},
  author={Richter, Stephan R and Vineet, Vibhav and Roth, Stefan and Koltun, Vladlen},
  booktitle={European Conference on Computer Vision},
  pages={102--118},
  year={2016},
  organization={Springer}
}

@inproceedings{Cauvis,
  title={Towards Single-Source Domain Generalized Object Detection via Causal Visual Prompts},
  author={Li, Chen and Xu, Huiying and Gao, Changxin and Wang, Zeyu and Liu, Yun and Zhu, Xinzhong},
  booktitle={The Thirty-ninth Annual Conference on Neural Information Processing Systems},
  year={2025}
}

@inproceedings{DINO,
  title={DINO: DETR with Improved DeNoising Anchor Boxes for End-to-End Object Detection},
  author={Zhang, Hao and Li, Feng and Liu, Shilong and Zhang, Lei and Su, Hang and Zhu, Jun and Ni, Lionel and Shum, Heung-Yeung},
  booktitle={The Eleventh International Conference on Learning Representations},
  year={2023}
}

@article{PGST,
  title={Phrase grounding-based style transfer for single-domain generalized object detection},
  author={Li, Hao and Wang, Wei and Wang, Cong and Wang, Mengzhu and Zhang, Xiang and Lan, Long and Liu, Xinwang and Li, Kenli and Cao, Xiaochun},
  journal={IEEE Transactions on Circuits and Systems for Video Technology},
  year={2025},
  publisher={IEEE}
}

@article{srcd,
  title={Srcd: Semantic reasoning with compound domains for single-domain generalized object detection},
  author={Rao, Zhijie and Guo, Jingcai and Tang, Luyao and Huang, Yue and Ding, Xinghao and Guo, Song},
  journal={IEEE Transactions on Neural Networks and Learning Systems},
  year={2024},
  volume={36},
  number={7},
  pages={12497-12506},
  publisher={IEEE}
}

@inproceedings{cityscapes,
  title={The cityscapes dataset for semantic urban scene understanding},
  author={Cordts, Marius and Omran, Mohamed and Ramos, Sebastian and Rehfeld, Timo and Enzweiler, Markus and Benenson, Rodrigo and Franke, Uwe and Roth, Stefan and Schiele, Bernt},
  booktitle={Proceedings of the IEEE Conference on Computer Vision and Pattern Recognition},
  pages={3213--3223},
  year={2016}
}

\textbf{Zihao Zhang}
received his M.Eng. degree from the College of Intelligence and Computing, Tianjin University in 2024. Currently, he is pursuing his Ph.D. degree at the College of Intelligence and Computing, Tianjin University. His research interests include computer vision and machine learning.

\textbf{Aming Wu}
received the Ph.D. degree from Tianjin University, Tianjin, China, in 2021. He is currently a Professor with the School of Computer Science and Information Engineering, Hefei University of Technology, Hefei, China. He has authored or coauthored more than 40 papers in top-tier journals and conferences. His research interests include computer vision, multimedia analysis, and machine learning.

\textbf{Yang Li} is currently pursuing the M.Eng. degree at the College of Intelligence and Computing, Tianjin University. His research interests include computer vision and machine learning, with a particular focus on visual representation learning and deep learning methods for real-world applications

\textbf{Yahong Han}
(Member, IEEE) received the Ph.D. degree from Zhejiang University, Hangzhou, China, in 2012. He is currently a professor with the College of Intelligence and Computing, Tianjin University, Tianjin, China. From Nov. 2014 to Nov. 2015, he visited Prof. Bin Yu’s group at UC Berkeley as a visiting scholar. His current research interests include multimedia analysis and computer vision.


\end{document}